\documentclass{article}

\usepackage{microtype}
\usepackage{graphicx}
\usepackage{booktabs} 
\usepackage{comment}
\usepackage{enumitem}

\usepackage[utf8]{inputenc} 
\usepackage[T1]{fontenc}    
\usepackage{hyperref}       
\usepackage{url}      
\usepackage{subfiles}
\usepackage{booktabs}       
\usepackage{amsfonts}       
\usepackage{nicefrac}  
\usepackage{xr}
\usepackage{microtype}      
\usepackage{amsmath, mathtools, amsthm, amssymb}
\usepackage{float}

\usepackage{tabularx}

\usepackage{algorithm}
\usepackage{algorithmic}

\usepackage{graphicx}
\usepackage{subfig}
\usepackage{hyperref}

\DeclareMathOperator{\EX}{\mathbb{E}}%

\usepackage{hyperref}

\usepackage[accepted]{icml2020}

\icmltitlerunning{oIRL: Robust Adversarial Inverse Reinforcement Learning with Temporally Extended Actions}

\begin{document}

\twocolumn[
\icmltitle{oIRL: Robust Adversarial Inverse Reinforcement Learning with Temporally Extended Actions}




\begin{icmlauthorlist}
\icmlauthor{David Venuto}{mcgill,mila}
\icmlauthor{Jhelum Chakravorty}{mcgill,mila}
\icmlauthor{Leonard Boussioux}{mit,mila}
\icmlauthor{Junhao Wang}{mcgill,mila}
\icmlauthor{Gavin McCracken}{mcgill,mila}
\icmlauthor{Doina Precup}{mcgill,mila,deepmind}
\end{icmlauthorlist}

\icmlaffiliation{mcgill}{Department of Computer Science, McGill University, Montreal, Canada}
\icmlaffiliation{mila}{Mila, Montreal, Canada}
\icmlaffiliation{mit}{Department of Operations Research, MIT, Cambridge, USA}
\icmlaffiliation{deepmind}{DeepMind, Montreal, Canada}

\icmlcorrespondingauthor{David Venuto}{david.venuto@mail.mcgill.com}
\icmlcorrespondingauthor{Doina Precup}{dprecup@cs.mcgill.ca}

\icmlkeywords{Machine Learning, ICML}

\vskip 0.3in
]



\printAffiliationsAndNotice{\icmlEqualContribution} 

\begin{abstract}
Explicit engineering of reward functions for given environments has been a major hindrance to reinforcement learning methods.  While Inverse Reinforcement Learning (IRL) is a solution to recover reward functions from demonstrations only, these learned rewards are generally heavily \textit{entangled} with the dynamics of the environment and therefore not portable or \emph{robust} to changing environments.  Modern adversarial methods have yielded some success in reducing reward entanglement in the IRL setting. In this work, we leverage one such method, Adversarial Inverse Reinforcement Learning (AIRL), to propose an  algorithm that learns hierarchical disentangled rewards with a policy over options. We show that this method has the ability to learn \emph{generalizable} policies and reward functions in complex transfer learning tasks, while yielding results in continuous control benchmarks that are comparable to those of the state-of-the-art methods. 
\end{abstract}

\section{Introduction}
\label{submission}

Reinforcement learning (RL) has been able to learn policies in complex environments but it usually requires designing suitable reward functions for successful learning. This can be difficult and may lead to learning sub-optimal policies with unsafe behavior \cite{45512} in the case of poor engineering. Inverse Reinforcement Learning (IRL) \cite{Ng00algorithmsfor,abbeel2004apprenticeship} can facilitate such reward engineering through learning an expert's reward function from expert demonstrations. 

IRL, however, comes with many difficulties and the problem is not well-defined because, for a given set of demonstrations, the number of optimal policies and corresponding rewards can be very large, especially for high dimensional complex tasks. Also, many IRL algorithms learn reward functions that are heavily shaped by environmental dynamics. Policies learned on such reward functions may not remain optimal with slight changes in the environment. Adversarial Inverse Reinforcement Learning (AIRL) \cite{airl} generates more \textit{generalizable} policies with \textit{disentangled reward} functions that are invariant to the environmental dynamics.  The reward and value function are learned simultaneously to compute the reward function in a state-only manner. This is an instance of a \textit{transfer learning problem with changing dynamics}, where the agent learns an optimal policy in one environment and then transfers it to an environment with different environmental dynamics. A practical example of this transfer learning problem would be teaching a robot to walk with some mechanical structure, and then generalize this knowledge to perform the task with different sized structural components.

Other methods have been developed to learn demonstrations in environments with differing dynamics and then exploit this knowledge while performing tasks in a novel environment. As complex tasks in different environments often come from several reward functions, methods such as Maximum Entropy IRL and GAN-Guided Cost Learning (GAN-GCL) tend to overgeneralize \cite{Finn:2016:GCL:3045390.3045397}.  One way to help solve the problem of over-fitting is to break down a policy into small \emph{option} (temporally extended action)-policies that solve various aspects of an overall task. This method has been shown to create a policy that is more generalizable \cite{Sutton:1999,10.5555/1577069.1755839}.  Methods such as Option-Critic have implemented modern RL architectures with a policy over options and have shown improvements for generalization of policies \cite{10.5555/3298483.3298491}. OptionGAN \cite{henderson2017optiongan} also proposed an IRL framework for a policy over options and is shown to have some improvement in one-shot transfer learning tasks, but does not return disentangled rewards. 

In this paper, we introduce Option-Inverse Reinforcement Learning (oIRL), to investigate transfer learning with options. Following AIRL, we propose an algorithm that computes disentangled rewards to learn joint reward-policy options, with each option-policy having rewards that are disentangled from environmental dynamics.  These policies are shown to be heavily portable in transfer learning tasks with differences in environments. We evaluate this method in a variety of continuous control tasks in the Open AI Gym environment using the MuJoCo simulator \cite{1606.01540,6386109} and GridWorlds with MiniGrid \cite{gym_minigrid}.  Our method shows improvements in terms of reward on a variety of transfer learning tasks while still performing better than benchmarks for standard continuous control tasks.
\section{Preliminaries}
\label{sec:prelim}

\textbf{Markov Decision Processes (MDP)} are defined by a tuple $\langle \mathcal{S}, \mathcal{A}, \mathbb P, R, \gamma \rangle$ where $\mathcal{S}$ is a set of states, $\mathcal{A}$ is the set of actions available to the agent, $\mathbb P$ is the transition kernel giving a probability over next states given the current state and action, $R : \mathcal{S} \times \mathcal{A} \rightarrow [0,R_{\text{max}}]$ is a reward function and $\gamma \in [0, 1]$ is a discount factor. $s_t$ and $a_t$ are respectively the state and action of the expert at time instant $t$. We define a policy $\pi$ as the probability distribution over actions conditioned on the current state; $\pi : S \times \mathcal{A} \rightarrow [0,1]$. A policy is modeled by a Gaussian distribution $\pi_\theta \sim \mathcal{N}(\mu,\,\sigma^{2})$ where $\theta$ is the policy parameters. The value of a policy is defined as $V_\pi(s) = \EX_\pi [\sum_{t=0}^{\infty} \gamma^t r_{t+1} | s]$, where $\mathbb E$ denotes the expectation. 
An agent follows a policy $\pi$ and receives reward from the environment. A state-action value function is $Q_\pi(s,a)=\EX_\pi[\sum_{t=0}^{\infty} \gamma^t r_{t+1} | s,a]$. The advantage is $A_\pi(s,a)=Q_\pi(s,a)-V_\pi(s)$. $r(s,a)$ represents a one-step reward.

\textbf{Options} ($\omega \in \Omega$) are defined as a triplet ($I_\omega,\pi_{\omega},\beta_{\omega}$), where $\pi_{\omega}$ is a policy over options, $I_{\omega} \in S$ is the initiation set of states and $\beta_{\omega} : S \rightarrow [0,1]$ is the termination function.  The policy over options is defined by $\pi_{\Omega}$. An option has a reward $r_{\omega}$ and an option policy $\pi_{\omega}$.

The policy over options is parameterized by $\zeta$, the intra-option policies by $\alpha$ for each option, the reward approximator by $\theta$, and the option termination probabilities by $\delta$.

In the one-step case, selecting an option using the policy-over-options can be viewed as a mixture of completely specialized experts. This overall policy can be defined as $\pi_{\Theta}(a|s)=\sum_{\omega \in \Omega}\pi_{\Omega}(\omega|s)\pi_{\omega}(a|s)$.

\textbf{Disentangled Rewards} are formally defined as a reward function $r_{\theta}^{*}(s, a, s')$ that is disentangled with respect to (w.r.t.) a ground-truth reward and a set of environmental dynamics $\mathcal{T}$ such that under all possible dynamics $T \in \mathcal{T}$, the optimal policy computed w.r.t. the reward function is the same.

\section{Related Work}
\label{sec:related}
\textbf{Generative Adversarial Networks (GANs)} learn the generator distribution $p_g$ and discriminator $D_{\theta_D}(\mathbf{x})$.  They use a prior distribution over input noise variables $p(\mathbf{z})$.  Given these input noise variables the mapping $G_{\theta_g}(\mathbf{z})$ is learned, which maps these input noise variables to the data set space. $G$ is a neural network.  Another neural network, $D_{\theta_D}(\mathbf{x})$, learns to estimate the probability that $\mathbf{x}$ came from the data set and not the generator $p_g$.

In our two-player adversarial training procedure, $D$ is trained to maximize the probability of assigning the correct labels to the dataset and the generated samples. $G$ is trained to minimize the objective $\text{log}(1-D_{\theta_D}(G_{\theta_G}(\mathbf{z})))$, which causes it to generate samples that are more likely to fool the discriminator. 

\textbf{Policy Gradient} methods optimize a parameterized policy $\pi_\theta$ using a gradient ascent.  Given a discounting term, the objective to be optimized is $p(\theta,s_0) = \EX[\sum_{t=0}^{\infty} \gamma^t r_{\theta}(s_t) | s_0]$. Proximal policy optimization (PPO) ~\citep{schulman2017proximal} is a policy gradient method that uses policy gradient theorem, which states $\frac{\partial p(\theta,s_0)}{\partial \theta}=\sum_{s}^{}\sum_{t=0}^{\infty}\gamma^t P(s_t=s|s_0)\sum_{a}^{}\frac{\partial \pi(a|s)}{\partial \theta}Q_{\pi_\theta}(s,a)$.  PPO has been adapted for the option-critic architecture (PPOC)~\citep{DBLP:journals/corr/abs-1712-00004}.

\textbf{Inverse Reinforcement Learning} (IRL) is a form of \emph{imitation learning}~\footnote{Here the agent learns the expert's policy by observing expert demonstrations~\citep{Ng00algorithmsfor}.}, where the expert's reward is estimated from demonstrations and then forward RL is applied to that estimated reward to find the optimal policy. Generative Adversarial Imitation Learning (GAIL) directly extracts optimal policies from expert's demonstrations ~\citep{ho_generative_2016}.  IRL infers a reward function from expert demonstrations, which is then used to optimize a generator policy.

In IRL, an agent observes a set of state-action trajectories from an expert demonstrator $\mathcal{D}$. We let $\mathcal{T}_\mathcal{D}=\{\tau^E_1,\tau^E_2,\dots,\tau^E_n\}$ be the state-action trajectories of the expert, $\tau^E_i \sim \tau_{\mathcal{D}}$ where $\tau^E_i = \{s_0,a_0,s_1,a_1\dots,s_k,a_k\}$. We wish to find the reward function $r(s,a)$ given the set of demonstrations $\mathcal{T}_\mathcal{D}$.  It is assumed that the demonstrations are drawn from the optimal policy $\pi^*(a|s)$. The Maximum Likelihood Estimation (MLE) objective in the IRL problems is therefore:
\begin{equation}
\text{}\max_\theta J(\theta)= \text{}\max_\theta  \EX_{\tau \sim \tau^E}[\log(p_\theta(\tau))],
\end{equation}
with $p_\theta(\tau) \propto p(s_0)\prod_{t=0}^T p(s_{t+1}|s_t,a_t)\exp{\left({\gamma^t r_{\theta}(s_t,a_t)}\right)}$.

\textbf{Adversarial Inverse Reinforcement Learning (AIRL)} is based on GAN-Guided Cost Learning \citep{gcl}, which casts the MLE objective as a Generative Adversarial Network (GAN) ~\citep{gan} optimization problem over trajectories.  In AIRL~\citep{airl}, the discriminator probability $D_{\theta}$ is evaluated using the state-action pairs from the generator (agent), as given by
\begin{equation}\label{eq:AIRL_D}
D_{\theta}(s,a) = \frac{\exp(f_{\theta}(s,a))}{\exp(f_{\theta}(s,a)) + \pi(a | s)}.
\end{equation}

The agent tries to maximize $R(s,a)=\log(1-D_{\theta}(s,a))-\log(D_{\theta}(s,a))$ where $f_\theta(s,a)$ is a learned function and $\pi$ is pre-computed.  This formula is similar to GAIL but with a \emph{recoverable reward function} since GAIL outputs 0.5 for the reward of all states and actions at optimality.  The discriminator function is then formulated as $f_{\theta,\Phi}(s,a,s') = g_{\theta}(s,a) + \gamma h_{\Phi}(s') - h_{\Phi}(s)$ given shaping function $h_{\Phi}$ and reward approximator $g_\theta$.  Under deterministic dynamics, it is shown in AIRL that there is a state-only reward approximator ( $f^*(s,a,s')=r^*(s)+\gamma V^*(s') - V^*(s) = A^*(s,a)$ where the reward is invariant to transition dynamics and is disentangled.

\textbf{Hierarchical Inverse Reinforcement Learning} learns policies with high level temporally extended actions using IRL. OptionGAN ~\citep{henderson2017optiongan} provides an adversarial IRL objective function for the discriminator with a policy over options.  It is formulated such that $L_{\text{reg}}$ defines the regularization terms on the mixture of experts so that they converge to options.  The discriminator objective in OptionGAN takes state-only input and is formulated as:
\begin{align}
L_{\Omega} &= \EX_{\omega}[\pi_{\Omega,\zeta}(\omega|s)(L_{\alpha,\omega})] + L_{reg}, \notag
\shortintertext{where} 
L_{\alpha,\omega} &= \EX_{\tau^N}[\log( r_{\theta,\omega}(s))] + \EX_{\tau^E}[\log(1- r_{\theta,\omega}(s))]. \label{eq:L}
\end{align}
In Directed-Info GAIL \citep{sharma2018directedinfo} implements GAIL in a policy over options framework.

Work such as \citep{Krishnan2016HIRLHI} solves this hierarchical problem of segmenting expert demonstration transitions by analyzing the changes in \emph{local linearity w.r.t a kernel function}.  It has been suggested that decomposing the reward function is not enough \citep{henderson2017optiongan}.  Other works have learned the latent dimension along with the policy for this task \citep{NIPS2017_6723, Wang:2017:RID:3295222.3295284}. In this formulation, the latent structure is encoded in an unsupervised manner so that the desired latent variable does not need to be provided.  This work parallels many hierarchical IRL methods but with recoverable robust rewards.

\section{MLE Objective for IRL Over Options}
\label{sec:result}
Let $(s_0,a_0, \dots s_T, a_T) \in \tau^{E}_{i}$ be an expert trajectory of state-action pairs.  Denote by $(s_0,a_0, \omega_0 \dots s_T, a_T, \omega_T) \in \tau_{\pi_{\Theta,t}}$ a novice trajectory generated by policy over options $\pi_{\Theta,t}$ of the generator at iteration $t$. 

Given a trajectory of state-action pairs, we first define an option transition probability given a state and an option.  Similar transition probabilities given state, action or option information are defined in (Appendix A).
\begin{align}
&P(s_{t+1}, \omega_{t+1}\,|\, s_t,\omega_t) \notag \\
& =\sum_{a \in A} \pi_{\omega,\alpha}(a | s_t) P(s_{t+1} | s_t,a)((1-\beta_{\omega_t,\delta}(s_{t+1}))\textbf{1}_{\omega_t=\omega_{t+1}} \notag \\
& \hskip 4em + \beta_{\omega_t,\delta}(s_{t+1})\pi_{\Omega,\zeta}(\omega_{t+1} | s_{t+1})). \label{eq:transition_augmented_state}
\end{align}

We can similarly define a discounted return recursively. Consider the policy over options based on the probabilities of terminating or continuing the option policies given a reward approximator $\hat{r}_{\theta}(s,a)$ for the state-action reward.
\begin{align}
&R^{}_{\theta,\delta}(s,\omega,a) \coloneqq \EX\Big[\hat{r}_{\theta,\omega}(s,a) + \gamma \sum_{s^{'} \in S}  P(s^{'}|s,a) \notag \\
& \Big(\beta_{\omega,\delta}(s^{'})R^{\Omega}_{\zeta,\theta,\alpha,\delta}(s^{'}) + \big(1-\beta_{\omega,\delta}(s^{'})\big)R_{\theta,\alpha,\delta}(s^{'},\omega)\Big)\Big]. \label{eq:dis_return}
\end{align}
$\omega_0$ is selected according to $\pi_{\zeta,\Omega}(\omega | s)$. The expressions for all relevant discounted returns appearing in the analysis are given in Appendix B. A suitable parameterization of the discounted return $R$ can be found by maximizing the causal entropy $\EX_{\tau \sim \mathcal{D}}[\log(p_{\theta}(\tau))]$ w.r.t parameter $\theta$. We then have for a trajectory $\tau$ with $T$ time-steps:
\begin{align}
& p_{\theta}(\tau) \\
&\approx p(s_0,\omega_0)\prod_{t=0}^{T-1} P(s_{t+1}, \omega_{t+1} | s_t, \omega_t, a_t)e^{R^{}_{\theta,\delta}(s_t,\omega_t,a_t)}. \notag
\end{align}

\subsection{MLE Derivative}
Similar to~\cite{airl} and \cite{gcl}, we define the MLE objective for the generator $p_\theta$ as 
\begin{align}
&J(\theta) = \EX_{\tau \sim \tau^E}[\sum_{t=0}^{T} R^{\Omega}_{\zeta,\theta,\delta}(s_t,a_t)] \notag \\
& \hskip 2em - \EX_{p_{\theta}}[\sum_{t=0}^{T} \sum_{\omega \in \Omega}\pi_{\zeta,\Omega}(\omega | s_t) R^{}_{\theta,\delta}(s_t,\omega,a_t)]. \label{eq:J_theta}
\end{align}
Note that we may or may not know the option trajectories in our expert demonstrations, instead they are estimated according to the policy over options.  The gradient of~\eqref{eq:J_theta} w.r.t $\theta$ (See Appendix B for detailed derivations) is given by:
\begin{align}
& \frac{\partial}{\partial \theta}J(\theta) =
\EX_{\tau \sim \tau^E}\Big[\frac{\partial}{\partial \theta}\log(p_\theta(\tau))\Big] \notag \\
& \approx \EX_{\tau \sim \tau^E}\Big[\sum_{t=0}^{T}  \frac{\partial}{\partial \theta}R^{\Omega}_{\zeta,\theta,\delta}(s_t,a_t)\Big]  \notag  -  \EX_{p_{\theta}}\Big[\sum_{t=0}^{T}  \frac{\partial}{\partial \theta} R^{\Omega}_{\zeta,\theta,\delta}(s_t,a_t)\Big]. \label{eq:J-derivative} 
\end{align}
We define $p_{\theta,t}(s_t,a_t)=\int_{s_{t^{'} \neq t}, a_{t^{'} \neq t}} p_{\theta}(\tau) ds_{t'} da_{t'}$ as the state action marginal at time $t$.  This allows us to  examine the trajectory from step $t$ as defined similarly in \citep{airl}. Consequently, we have
\begin{align}
\frac{\partial}{\partial \theta}J(\theta) &= \sum_{t=0}^{T} \EX_{\tau \sim \tau^E}\biggl[\frac{\partial}{\partial \theta}R^{\Omega}_{\zeta,\theta,\delta}(s_t,a_t)\biggr] \tag{xyz}  \\
& \hskip 3em - \EX_{p_{\theta,t}}\biggl[\frac{\partial}{\partial \theta}R^{\Omega}_{\zeta,\theta,\delta}(s_t,a_t)\biggr].
\label{eq:derivative_J}
\end{align}

Since $p_{\theta}$ is difficult to draw samples from, we estimate it using importance sampling distribution over the generator density. Then, we compute an importance sampling estimate of a \emph{mixture policy} $\mu_{t,w}(\tau)$ for each option $w$ as follows.  

We sample a mixture policy $\mu_{\omega}(a|s)$ defined as $\frac{1}{2} \pi_{\omega}(a|s) + \frac{1}{2}\hat{p}_{\omega}(a | s)$ and $\hat{p}_{\omega}(a|s)$ is a density estimate trained on the demonstrations.  We wish to minimize $D_{KL}(\pi_w(\tau) | p_{\omega}(\tau))$ to reduce the importance sampling distribution variance, where $D_{KL}$ is the \emph{Kullback–Leibler divergence} metric~\cite{KL_original_paper}  between two probability distributions. Applying the aforementioned density estimates in~\eqref{eq:derivative_J}, we can express the gradient of the MLE objective $J$ follows: 
\begin{align}
& \frac{\partial}{\partial \theta}J(\theta) = \sum_{t=0}^{T} \EX_{\tau \sim \tau^E}[\frac{\partial}{\partial \theta}R^{\Omega}_{\zeta,\theta,\delta}(s_t,a_t)] - \notag \\ 
& \hskip -2em \EX_{\mu_{t}}\biggl[\sum_{\omega \in \Omega}\pi_{\zeta,\Omega}(\omega | s_t)\frac{p_{\theta,t,\omega}(s_t,a_t)}{\mu_{t,w}(s_t,a_t)}
\frac{\partial}{\partial \theta}R^{}_{\theta,\delta}(s_t,\omega,a_t)\biggr],\label{eq:mod_MLE_grad}
\end{align}
where
\begin{align}
&\frac{\partial}{\partial \theta}R^{\Omega}_{\zeta,\theta,\alpha,\delta}(s) = \EX\Bigg[\sum_{\omega \in \Omega}\pi_{\Omega,\zeta}(\omega|s)\Big[\sum_{a \in A}\pi_{\omega,\alpha}(a|s) \notag \\ 
& \Big(\frac{\partial}{\partial \theta}\hat{r}_{\theta}(s,a) + \gamma \sum_{s^{'} \in S} P(s^{'}|s,a)\Big(\beta_{w,\delta}(s^{'})\frac{\partial}{\partial \theta}R^{\Omega}_{\zeta,\theta,\alpha,\delta}(s^{'}) \notag \\
& \hskip 3em + (1-\beta_{\omega,\delta}(s^{'}))\frac{\partial}{\partial \theta}R^{}_{\theta,\alpha,\delta}(s^{'},\omega)\Big)\Big)\Big]\Bigg]. \label{eq:R_Omega_derivative}
\end{align}

\section{Discriminator Objective}
In this section we formulate the discriminator, parameterized by $\theta$, as the odds ratio between the policy and the exponentiated reward distribution for option $\omega$.  We have a discriminator $D_{\theta,\omega}$ for each option $\omega$ and a sample generator option policy $\pi_w$, defined as follows:
\begin{equation}
D_{\theta,\omega} (s,a) = \frac{\exp(f_{\theta,\omega}(s,a))}{\exp(f_{\theta,\omega}(s,a)) + \pi_w(a | s)}.
\end{equation}

\subsection{Recursive Loss Formulation}

The discriminator $D_{\theta,\omega}$ is trained by minimizing the cross-entropy loss between expert demonstrations and generated examples assuming we have the same number of options in the generated and expert trajectories.
We define the loss function $L_\theta$ as follows: 
\begin{align}
& l_{\theta}(s,a,\omega) \label{eq:L_theta} \\
& = -\EX_{\mathcal{D}}[\log(D_{\theta,\omega}(s,a))]-
\EX_{\pi_{\Theta,t}}[\log(1-D_{\theta,\omega}(s,a))]. \notag
\end{align}
The parameterized total loss for the entire trajectory, $_{L\theta,\alpha,\delta}(s,a,\omega)$, can be expressed recursively as follows by taking expectations over the next options and states:
\begin{align} 
&L^{}_{\theta,\delta}(s,a,\omega) \notag \\
&= l_{\theta}(s,a,\omega) +
\gamma \sum_{s^{'} \in S}  P(s^{'}|s,a)\Big(\beta_{w,\delta}(s^{'})L^{\Omega}_{\zeta,\theta,\alpha,\delta}(s^{'}) \notag \\
& \hskip 7em + \big(1-\beta_{w,\delta}(s^{'})\big)L^{}_{\theta,\alpha,\delta}(s^{'},w)\Big) \label{eq:L_total} \\  
& L_{\theta,\alpha,\delta}(s,w) \coloneqq \EX_{a \in A}[L^{}_{\theta,\delta}(s,w,a)] \label{eq:L}\\ 
& L^{\Omega}_{\zeta,\theta,\delta}(s,a) \coloneqq \EX_{w \in \Omega}[L^{}_{\theta,\delta}(s,w,a)] \label{eq:L_Omega_1}\\
&L^{\Omega}_{\zeta,\theta,\alpha,\delta}(s) \coloneqq \EX_{\omega \in \Omega}[L_{\theta,\alpha,\delta}(s,\omega)]. \label{eq:L_Omega_2}
\end{align}
The agent wishes to minimize $L^{}_{\theta,\alpha,\delta}$ to find its optimal policy.

\subsection{Optimization Criteria}
For a given option $\omega$, define the reward function $\hat{R^{}}_{\theta,\delta}(s,\omega,a)$, which is to be maximised. We then write a  negative discriminator loss ($-L^{D}$) to turn our loss minimization problem into a maximization problem, as follows:
\begin{align}
&\hskip -2em -L^D=\hat R_{\theta,\delta}(s,\omega,a) = \notag \\ 
&\log( D_{\theta,\omega}(s,a)) - \log(1- D_{\theta,\omega}(s,a)). \label{eq:L_D}
\end{align}
We set a mixture of experts and novice as $\bar{\mu}$ observations in our gradient.   We then wish to take the derivative of the inverse discriminator loss as,
\begin{equation}
\begin{aligned}
& \frac{\partial}{\partial \theta} (-L^{D}) = \sum_{t=0}^{T} \EX_{\tau \sim \tau^E}\left[\sum_{\omega \in \Omega}\pi_{\Omega,\zeta}(\omega|s_t)\frac{\partial}{\partial \theta}\Big(-L^{}_{\theta,\delta}(s_t,\omega,a_t)\Big)\right]-\\ & \EX_{\bar{\mu}_t}\Big[\sum_{\omega \in \Omega}\pi_{\Omega,\zeta}(\omega|s_t) \left(\frac{\exp(-L^{}_{\theta,\delta}(s_t,\omega,a_t))}{\frac{1}{2}\exp(-L^{}_{\theta,\delta}(s_t,\omega,a_t)) +\frac{1}{2} \pi_{\omega}(a_t | s_t)}\right) \\
& \frac{\partial}{\partial \theta}\Big(-L^{}_{\theta,\delta}(s_t,\omega,a_t)\Big)\Big].
\end{aligned}
\end{equation}
We can multiply the top and bottom of the fraction in the mixture expectation by the state marginal $\pi_{\omega}(s_t) = \int_{a \in A} \pi_{\omega}(s_t,a_t)$.  This allows us to write  $\hat{p}_{\theta,t,\omega}(s_t,a_t) = \exp(L_{\theta,\delta}(s_t,\omega,a_t))\pi_{\omega,t}(s_t)$. Using this, we can derive an importance sampling distribution in our loss,
\begin{equation}
 \label{eq:derivative_L}
\begin{aligned}
& \frac{\partial}{\partial \theta} (-L^D) = \sum_{t=0}^{T} \EX_{\tau \sim \tau^E}\left[\sum_{\omega \in \Omega}\pi_{\Omega,\zeta}(\omega|s_t)\frac{\partial}{\partial \theta}\Big(-L^{}_{\theta,\delta}(s_t,\omega,a_t)\Big)\right]\\ & - \EX_{\bar{\mu}_t}\left[\sum_{\omega \in \Omega}\pi_{\Omega,\zeta}(\omega|s_t) \left(\frac{\hat{p}_{\theta,t,\omega}(s_t,a_t)}{\hat{\mu}_{t,\omega}(s_t,a_t)}\right)\frac{\partial}{\partial \theta}\Big(-L^{}_{\theta,\delta}(s_t,\omega,a_t)\Big)\right].  \\
\end{aligned}
\end{equation}
The gradient of this parametrized reward function corresponds to the inverse of the discriminator's objective:
\begin{align}
&\frac{\partial}{\partial \theta} \hat{R^{}}_{\theta,\delta}(s,\omega,a) \approx \frac{\partial}{\partial \theta} \Big(-L^{}_{\theta,\delta}(s,\omega,a)\Big) \label{eq:derivative-R-hat} \notag\\
& \hskip -1em = \EX \Big[\frac{\partial}{\partial \theta}r_{\theta,\omega}(s,a)  + \gamma \sum_{s^{'} \in S}  P(s^{'}|s,a)\Big(\beta_{\omega,\delta}(s^{'}) \notag \\
& \frac{\partial}{  \partial \theta}\big(-L^{\Omega}_{\zeta,\theta,\alpha,\delta}(s^{'})\big) + \big(1-\beta_{\omega,\delta}(s^{'}) \big)\frac{\partial}{\partial \theta}\big(-L^{}_{\theta,\delta}(s^{'},\omega)\big)\Big)\Big].
\end{align}
See Appendix C for the  detailed derivations of the terms appearing in~\eqref{eq:derivative-R-hat}. Substituting~\eqref{eq:derivative-R-hat} into~\eqref{eq:derivative_L} one can   see that~\eqref{eq:mod_MLE_grad} (derivative of MLE objective)  and~\eqref{eq:R_Omega_derivative} are of the same form as of~\eqref{eq:derivative_L} (derivative of the discriminator objective and~\eqref{eq:derivative-R-hat}).

\section{Learning Disentangled State-only Rewards with Options}
\label{disnt}
In this section, we provide our main algorithm for learning robust rewards with options. Similar to AIRL, we implement our algorithm with a discriminator update that considers the rollouts of a policy over options.  We perform this update with $(s,a,s')$ triplets and a discriminator function in the form of $f_{\theta,\omega}(s,a,s')$ as given in~\eqref{eq:SA_only}. This allows us to formulate the discriminator with state-only rewards in terms of option-value function estimates.  We can then compute an option-advantage estimate.  Since the reward function only requires state, we learn a reward function and corresponding policy that is disentangled from the environmental transition dynamics.
\begin{equation}
\label{eq:SA_only} f_{\omega,\theta}(s,a,s') = \hat{r}_{\omega,\theta}(s) + \gamma \hat{V}_{\Omega}(s') - \hat{V}_{\Omega}(s) = \hat{A}(s,a,\omega)
\end{equation}
Where $Q(s,\omega)=\sum_{a \in \mathcal{A}}\pi_{\omega,\alpha}(a|s)[r_{\omega,\theta}(s,a)+\gamma \sum_{s' \in \mathcal{S}}P(s'|s,a) ((1-\beta_{\delta,\omega}(s'))Q(s',\omega) +\beta_{\delta,\omega}(s')V_{\Omega}(s'))]$ and $V_{\Omega}(s)=\sum_{\omega \in \Omega}\pi_{\Omega,\zeta}(\omega|s)Q(s,\omega)$.

Our discriminator model must learn a parameterization of the reward function and the value function for each option, given the total loss function in~\eqref{eq:L}.  These parameterized models are learned with a multi-layer perceptron. For each option, the termination functions $\beta_{\omega,\delta}$ and option-policies $\pi_{\omega,\alpha}$ are learned using PPOC.

\subsection{The main algorithm: Option-Adversarial Inverse Reinforcement Learning (oIRL)}


Our main algorithm, oIRL,  is given by Algorithm~\ref{algo:IRL-robust}.  Here, we iteratively train a discriminator from expert and novice sampled trajectories using the derived discriminator objective.  This allows us to obtain reward function estimates for each option.  We then use any policy optimization method for a policy over options given these estimated rewards.

We can also have discriminator input of state-only format as described in~\eqref{eq:SA_only}. It is important to note that in our recursive loss, we recursively simulate a trajectory to compute the loss a finite number of times (and return if the state is terminal). We show the adversarial architecture of this algorithm in Appendix D.  

\begin{algorithm*}[]
\caption{IRL Over Options with Robust Rewards (oIRL)}\label{algo:IRL-robust}
 \begin{algorithmic}[1]
\REQUIRE $\text{Expert Trajectories: } \{\tau_1^E,\dots,\tau_n^E\}\in \mathcal{T}_\mathcal{D}, \text{Initial Parameters: }(\theta_0,\zeta_0,\delta_0,\alpha_0), \gamma$
\STATE $\text{Initialize policies } \pi_{\omega,\alpha_0}, \pi_{\Omega,\zeta_0} \text{ and discriminators } D_{\theta_0,\omega}, \text{ and } \beta_{\omega,\delta_0} \forall \omega \in \Omega$
\FOR{$\text{step } t=0,1,2,\dots, T$}
\STATE $\text{Collect trajectories }\tau_i = (s_0,a_0,\omega_0,\dots) \text{ from } \pi_{\omega,\alpha_t}, \pi_{\Omega,\zeta_t},\beta_{\omega,\delta_t}$
\STATE $\textbf{Train discriminator } D_{\theta_t,\omega}$
\FOR{$\text{step } k=0,1,2,\dots$}
\STATE $\text{Sample } (s_k,a_k,{s'}_k,\omega_k) \sim \tau_{i,t}$
\IF{$s'$ not terminal state}
\STATE $\text{Sample } \omega_{k}^{'} \sim \pi_{\Omega,\zeta_t}(\omega|s_{k}^{'}), a_{k,1}^{'} \sim \pi_{\omega_{k}^{'},\alpha_{t}}(a|s_{k}^{'}), a_{k,2}^{'} \sim \pi_{\omega_{k},\alpha_{t}}(a|s_{k}^{'})$
\STATE $\text{ Observe } s^{''}_{k,1},s^{''}_{k,2} \text{ from environment}$ 
\STATE $L_{k}(s_k,a_k,s_{k}^{'},\omega_k) = -\EX_{\mathcal{D}}[\log(D_{\theta_t,\omega_k}(s_k,a_k,s_{k}^{'}))]-\EX_{\pi_{\Theta,t}}[\log(1-D_{\theta_t,\omega_k}(s_k,a_k,s_{k}^{'}))]$
\STATE $\text{Optimize model parameters w.r.t.: } -L^D = L_k + \gamma( \beta_{\delta_t,\omega_k}(s')L(s_{k}^{'},a_{k,1}^{'},s_{k,1}^{''},\omega_{k}^{'})$
\STATE $+(1-\beta_{\delta_t,\omega_k}(s_k^{'}))L(s_{k}^{'},a_{k,2}^{'},s_{k,2}^{''},\omega_{k}))$
\ENDIF
\ENDFOR
\STATE $\text{Obtain reward } r_{\theta_t,\omega}(s,a,s') \leftarrow \log( D_{\theta_t,\omega}(s,a,s')) - \log(1-D_{\theta_t,\omega}(s,a,s'))$
\STATE $\text{Update } \pi_{\omega,\alpha_t}, \beta_{\omega,\delta_t} \forall \omega \in \Omega \text{ and } \pi_{\Omega,\zeta_t} \text{ with any policy optimization method (e.g. PPOC)}$
\ENDFOR
\end{algorithmic}
\end{algorithm*}

\section{Convergence Analysis}
In this section we explain the gist of the analysis of convergence of oIRL. The detailed proofs can be found in Appendix E and F.

We first show that the actual reward function is recovered (up to a constant) by the reward estimators. We show that for each option's reward estimator $g_{\theta,\omega}(s)$,  we have $g^*_{\omega}(s) = r^*(s) + c_{\omega}$, where $c_{\omega}$ is a finite constant.  Using the fact that $g_{\theta,\omega}(s) \rightarrow  g^*_{\omega}(s) = r^*(s) + c_{\omega}$, and by using Cauchy-Schwarz inequality of sup-norm, we prove that the update of the TD-error is a contraction, i.e., 
\begin{equation}
    \max_{s'',\omega''}|Q_{\pi_{\Omega,t}}(s,\omega)-Q^*(s,\omega)| \leq \epsilon + \max_{\omega \in \Omega} c_\omega.
\end{equation}
In order to prove asymptotic convergence to the optimal option-value  $Q^*$, we show using the contraction argument that $g_{\theta,\omega}(s) + \gamma Q(s',\omega)$ converges to $Q^*$ by establishing the following inequality:
\begin{align}
    & |\mathbb{E}[g_{\theta,\omega}(s)] + \gamma \mathbb{E}[Q(s',\omega)|s] - Q^*(s',\omega)| \notag \\
    & \leq (\max_{\omega \in \Omega} c_\omega)(\epsilon+\max_{\omega \in \Omega} c_\omega) \gamma.
\end{align}
\section{Experiments}
oIRL learns disentangled reward functions for each option policy, which  facilitates policy generalizability and is instrumental in  \emph{transfer learning}.

Transfer learning can be described as using information learned by solving one problem and then applying it to a different but related problem. In the RL sense, it means taking a policy trained on one environment and then using the policy to solve a similar task in a different previously unseen environment. 

We run experiments in different environments to address the following questions:
\begin{itemize}
    \item Does learning a policy over options with the AIRL framework improve policy generalization and reward robustness in transfer learning tasks where the environmental dynamics are manipulated?
    \item Can the policy over options framework match or exceed benchmarks for imitation learning on complex continuous control tasks?
\end{itemize}
To answer these questions, we compare our model against AIRL (the current state of the art for transfer learning) in a transfer task by learning in an \emph{ant} environment and modifying the physical structure of the ant and compare our method on various benchmark IRL continuous control tasks. We wish to see if learning disentangled rewards for sub-tasks through the options framework is more portable.  

We train a policy using each of the baseline methods and our method on these expert demonstrations for 500 time steps on the \emph{gait} environments and 500 time steps on the hierarchical ones.  Then we take the trained policy (the parameterized distribution) and use this policy on the transfer environments and observe the reward obtained.  Such a method of transferring the policy is called a \emph{direct policy transfer}.

\subsection{Gait Transfer Learning Environments}
\begin{figure}[H]%
\begin{center}

    \subfloat[Ant environment ]{\includegraphics[height=2.4cm]{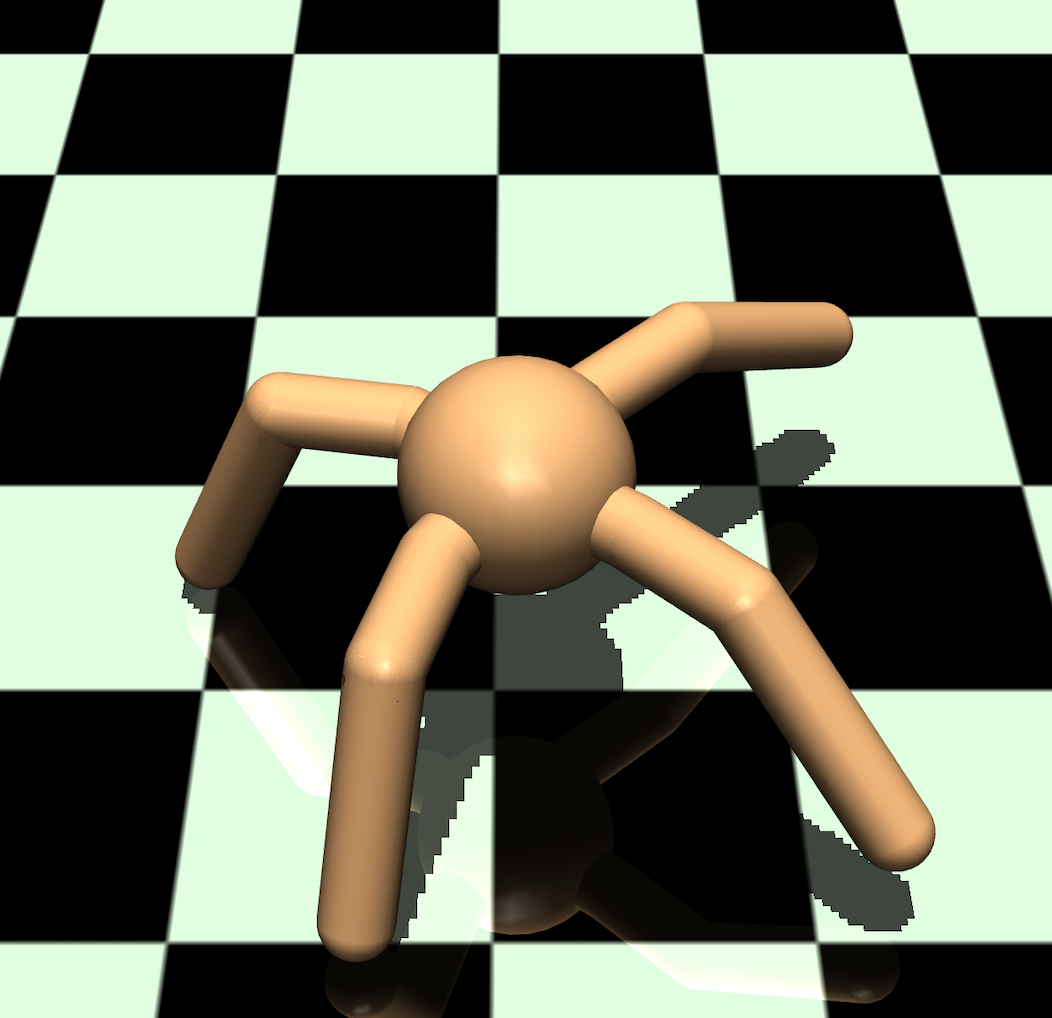}} \hspace{0.2cm}%
    \subfloat[Big Ant environment ]{\includegraphics[height=2.4cm]{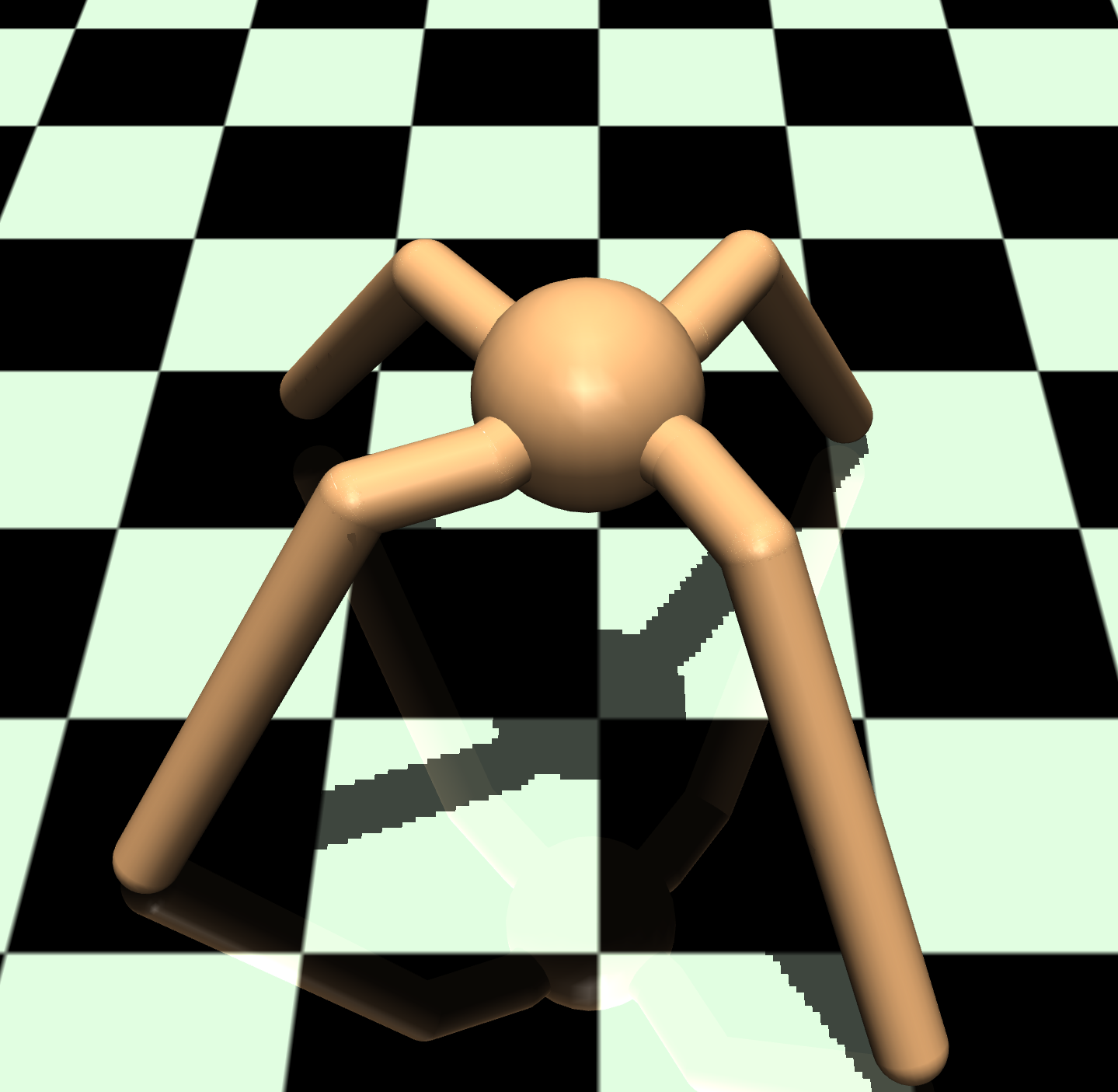}} \hspace{0.2cm}%
        \subfloat[Amputated Ant environment ]{\includegraphics[height=2.4cm]{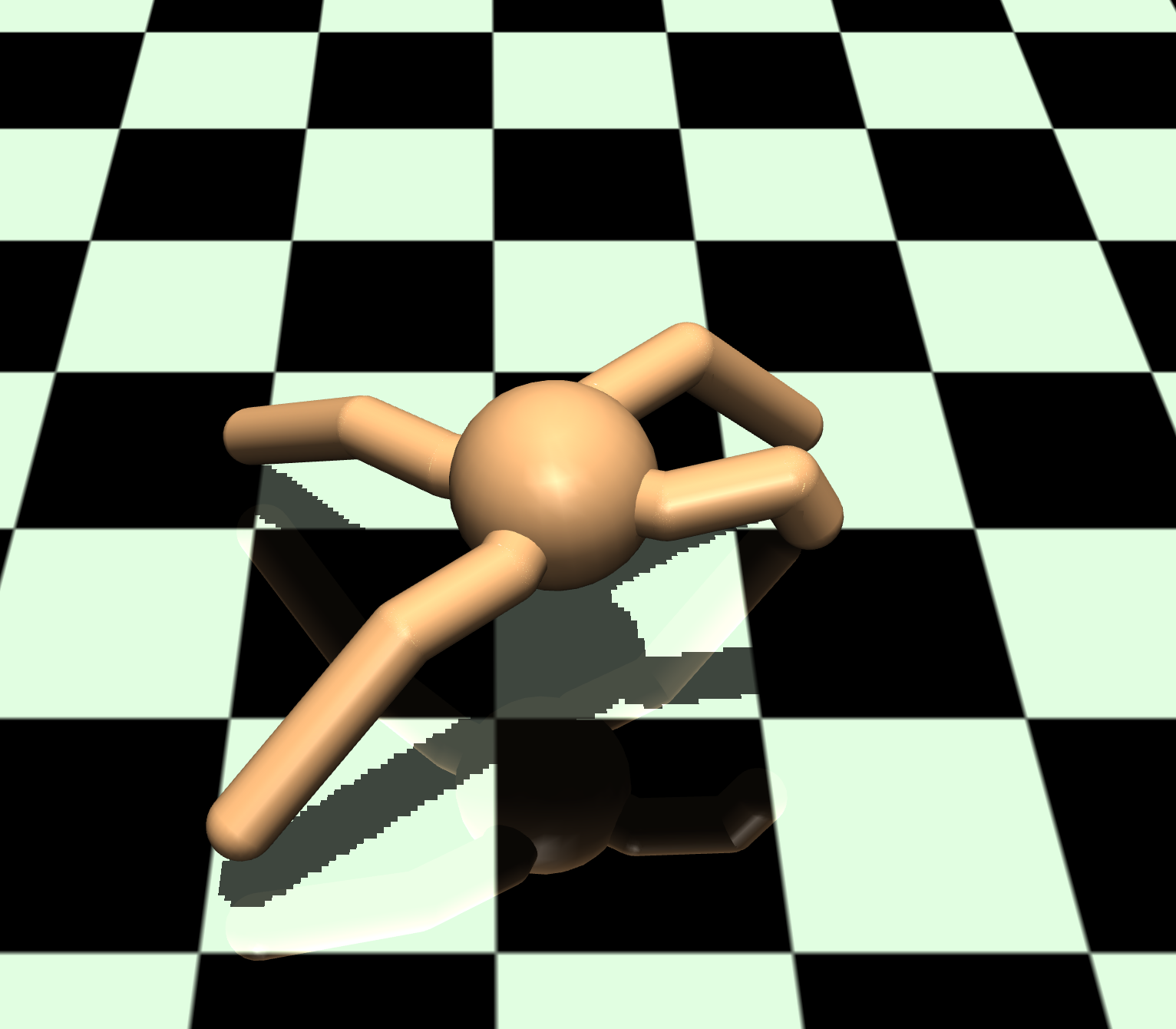}}%

        \caption{MuJoCo Ant Gait transfer learning task environments.  When the ant is disabled, it must position itself correctly to crawl forward.  This requires a different initial policy than the original environment where the ant must only crawl sideways.}
            \label{fig:world_envs}

        \end{center}
        \vskip -0.2in

\end{figure}

For the transfer learning tasks we use \emph{Transfer Environments for MuJoCo} \cite{transfer_git}, a set of gym environments for studying potential improvements in transfer learning tasks. The task involves an \emph{Ant} as an agent which optimizes a gait to crawl sideways across the landscape.  The expert demonstrations are obtained from the optimal policy in the basic Ant environment.  We disable the agent ant in two ways for two transfer learning tasks. In \emph{BigAnt} tasks, the length of all legs is doubled, no extra joints are added though. The \emph{Amputated Ant} task modifies the agent by shortening a single leg to disable it. These transfer tasks require the learning of a true disentangled reward of walking sideways instead of directly imitating and learning the reward specific to the gait movements.  These manipulations are shown in Figure \ref{fig:world_envs}.
Table \ref{gaitResults} shows the results in terms of reward achieved for the ant gait transfer tasks.  As we can see, in both experiments our algorithm performs better than AIRL. Remark that the ground truth is obtained with PPO after 2 million iterations (therefore much less sample efficient than IRL).
\begin{table}[t]
\caption{The mean reward obtained (higher is better) over 100 runs for the Gait transfer learning tasks. We also show the results of PPO optimizing the ground truth reward.}
\label{gaitResults}
\vskip 0.15in
\begin{center}
\begin{small}
\begin{sc}
\begin{tabular}{lcccr}
\toprule
& Big Ant & Amputated Ant \\
\midrule
    AIRL (Primitive)   & -11.6  & 134.3     \\
    2 Options oIRL    & \textbf{4.7}    & 122.6    \\
    4 Options oIRL        & -1.7 & \textbf{167.1} \\
    \textbf{Ground Truth} & 142.9  & 335.4
\end{tabular}
\end{sc}
\end{small}
\end{center}
\vskip -0.1in
\end{table}

\subsection{Maze Transfer Learning Tasks}

\begin{figure}[h]

\begin{center}

\subfloat[width=\columnwidth][LavaCrossing-M MiniGrid Env]{{\includegraphics[height=2.5cm]{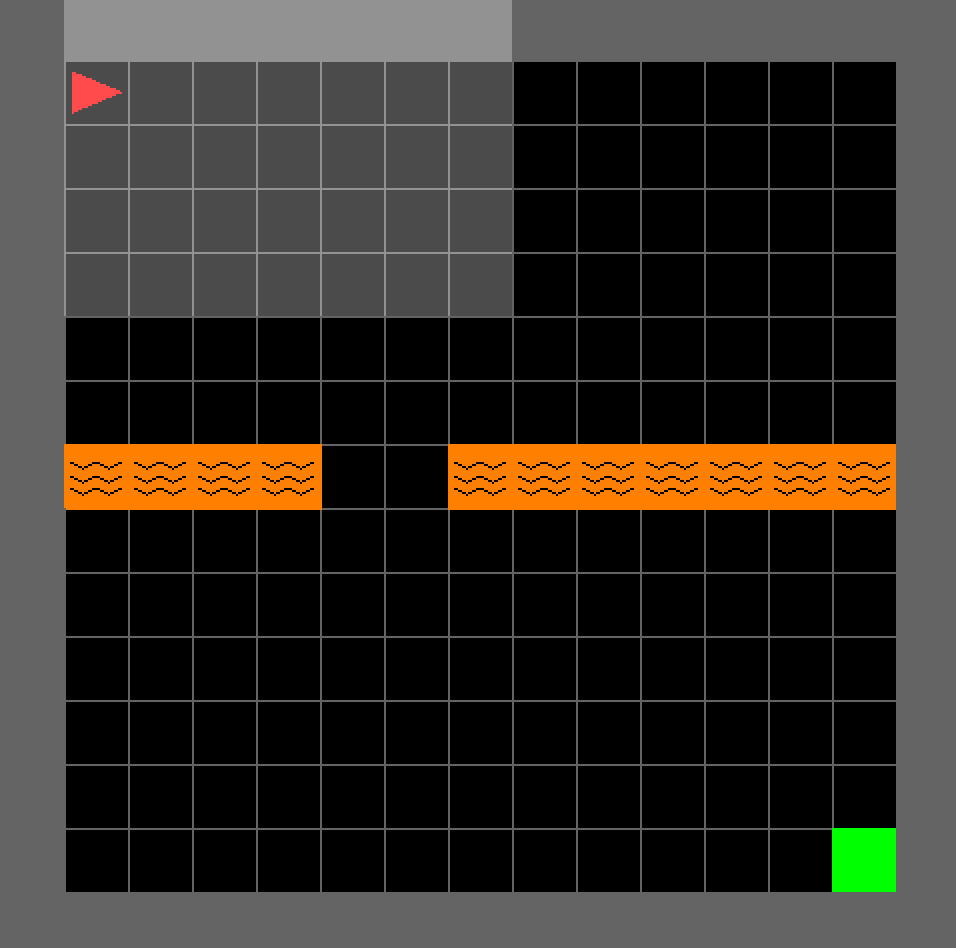}}}\hspace{0.5cm}%
    \subfloat[LavaCrossing-R MiniGrid Env]{{\includegraphics[height=2.5cm]{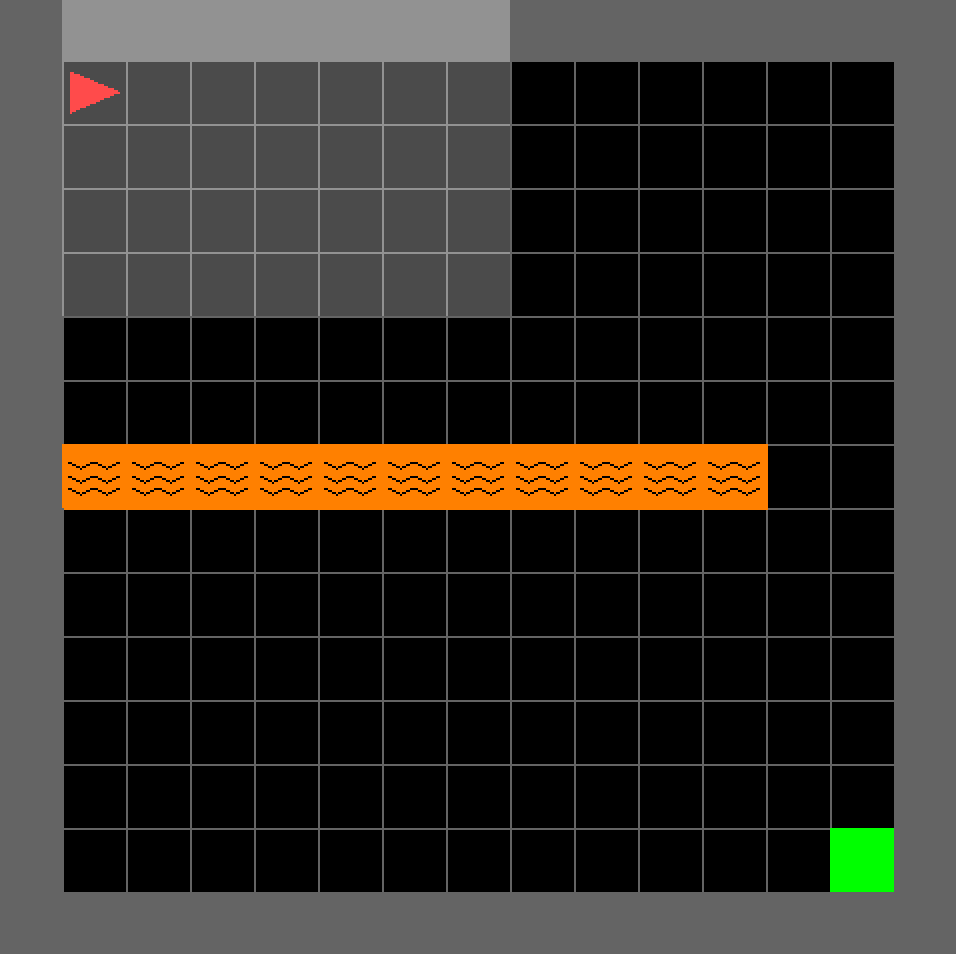}}} \hfill \\ \subfloat[FlowerMaze-R MiniGrid Env] {{\includegraphics[height=2.5cm]{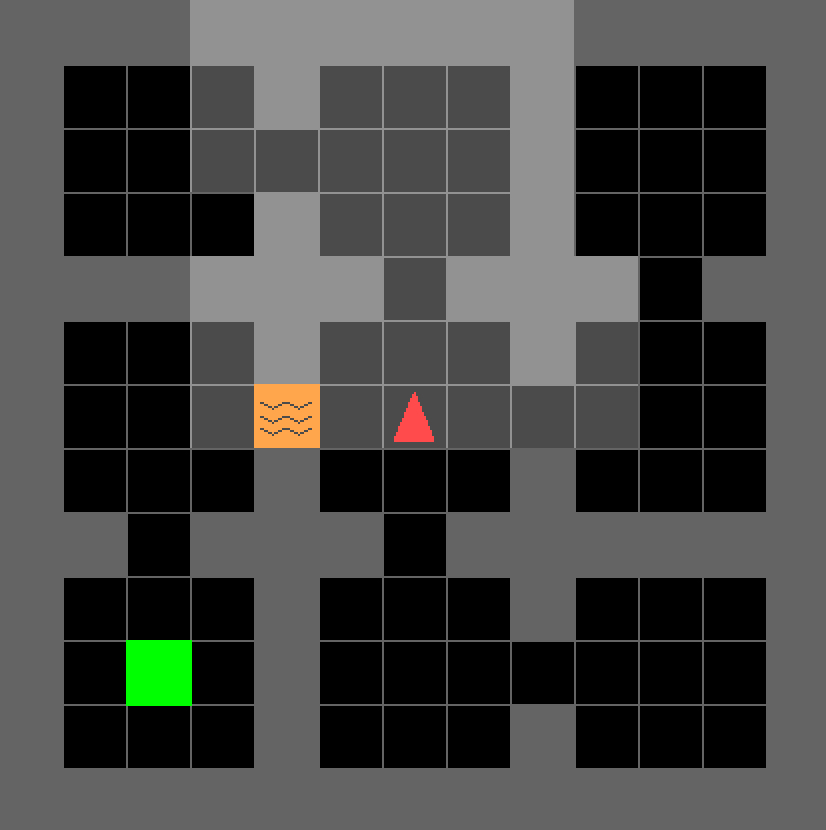}}}\hspace{0.5cm}%
    \subfloat[FlowerMaze-T MiniGrid Env]{{\includegraphics[height=2.5cm]{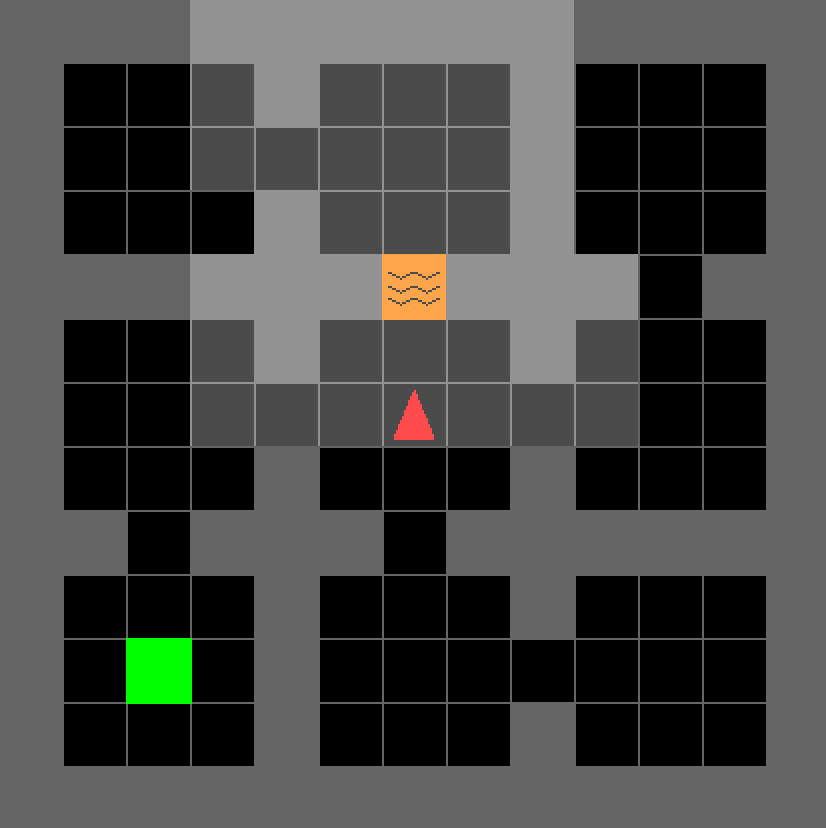}}}%
    \caption{The MiniGrid transfer learning task set 1.  Here the policy is trained on (a or c) using our method and the baseline methods and then transferred to be used on environment (b or d).  The green cell is the goal.}%
        \label{fig:transfer-grid}

    \end{center}

\vskip -0.2in
\end{figure}

We also create transfer learning environments in a 2D Maze environment with lava blockades.  The goal of the agent is to go through the opening in a row of lava cells and reach a goal on the other end.  For the transfer learning task, we train the agent on an environment where the "crossing" path requires the agent to go through the middle for (\emph{LavaCrossing-M}) and then the policy is directly transferred and used on a GridWorld of the same size where the crossing is on the right end of the room (\emph{LavaCrossing-R}).  An additional task would be changing a blockade in a \emph{Maze} (\emph{FlowerMaze-(R,T)}).  The two environments are shown in Figure \ref{fig:transfer-grid}.  We can think of two sub-tasks in this environment, \emph{going to the lava crossing} and then \emph{going to the goal}. 

In all of these environments, the rewards are sparse. The agent receives a non-zero reward only after completing the mission, and the magnitude of the reward is $1 - 0.9 \cdot n / n_{\text{max}}$, where $n$ is the length of the successful episode and $n_{\text{max}}$ is the maximum number of steps that we allowed for completing the episode, different for each mission.

We show the mean reward after 10 runs using the direct policy transfers on the environments in Table \ref{gridresults}.  The 4 option oIRL achieved the highest reward on the LavaCrossing tasks.  The FlowerMaze task was quite difficult with most algorithms obtaining very low reward.  Options still result in a large improvement. 

\begin{table}[t]
\caption{The mean reward obtained (higher is better) over 10 runs for the Maze transfer learning tasks. We also show the results of PPO optimizing the ground truth reward.}
\label{gridresults}
\vskip 0.15in
\begin{center}
\begin{small}
\begin{sc}
\begin{tabular}{lcccr}
\toprule
    & LavaCrossing & FlowerMaze\\
\midrule
    AIRL (Primitive) & 0.64  & 0.11      \\
    2 Options oIRL    & 0.67  & 0.20    \\
    4 Options oIRL    & \textbf{0.81}     &  \textbf{0.23} \\
    \textbf{Ground Truth} & 1.00  & 1.00
\end{tabular}
\end{sc}
\end{small}
\end{center}
\vskip -0.1in
\end{table}

\subsection{Hierarchical Transfer Learning Tasks}

In addition, we adopt more complex hierarchical environments that require both locomotion and object interaction. In the first environment, the ant must interact with a large movable block.  This is called the \emph{Ant-Push} environment \cite{10.5555/3045390.3045531}. To reach the goal, the ant must complete two successive processes: first, it must move to the left of the block and then push the block right, which clears the path towards the target location. There is a maximum of 500 timesteps. These can be thought of as hierarchical tasks with \emph{pushing to the left}, \emph{pushing to the right} and \emph{going to the goal} as sub-goals. 

\begin{figure}[h]

\begin{center}

\subfloat[Ant-Maze environment]{{\includegraphics[height=2.5cm]{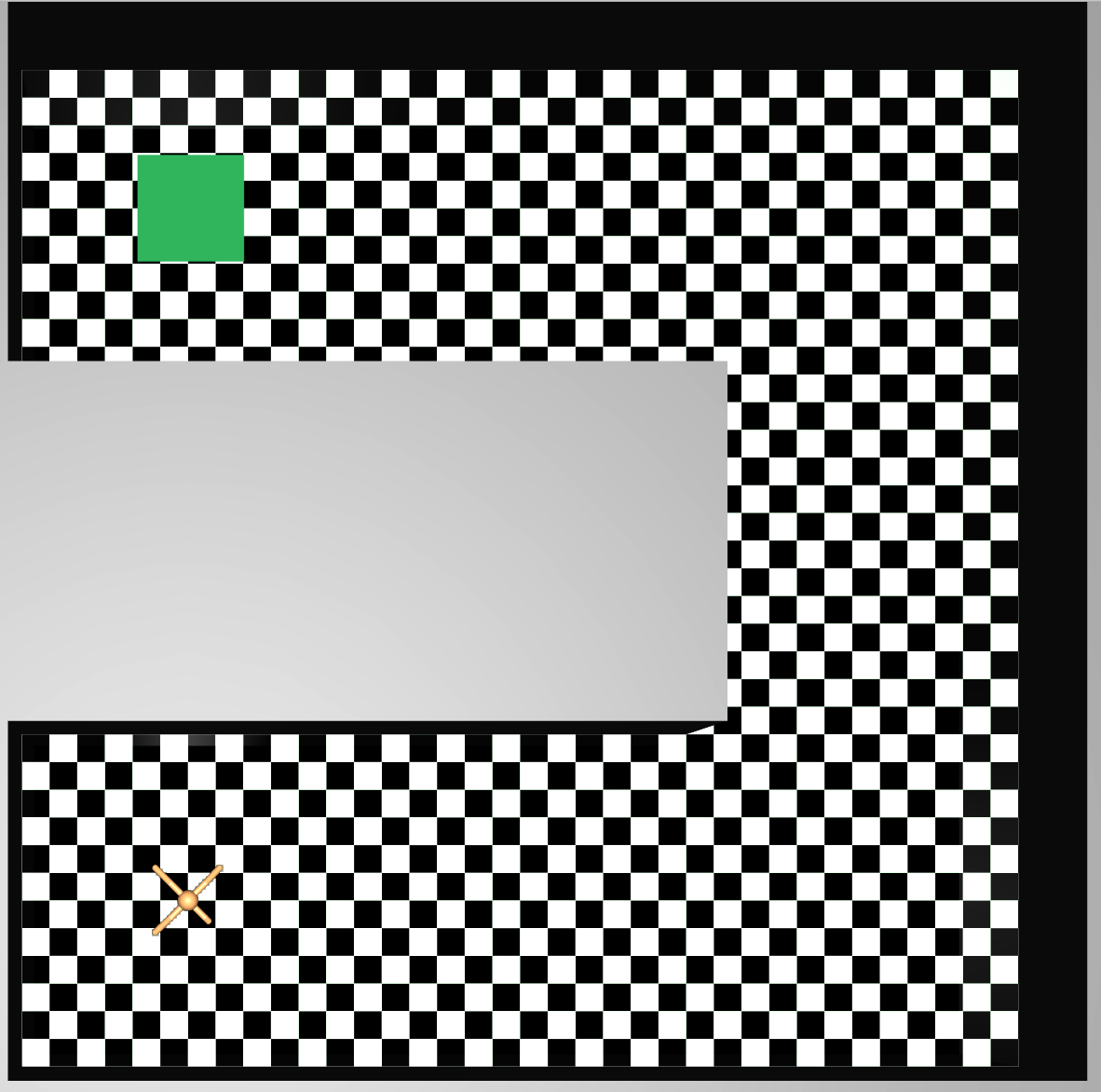}}}\hspace{0.5cm}
    \subfloat[Ant-Push environment]{{\includegraphics[height=2.5cm]{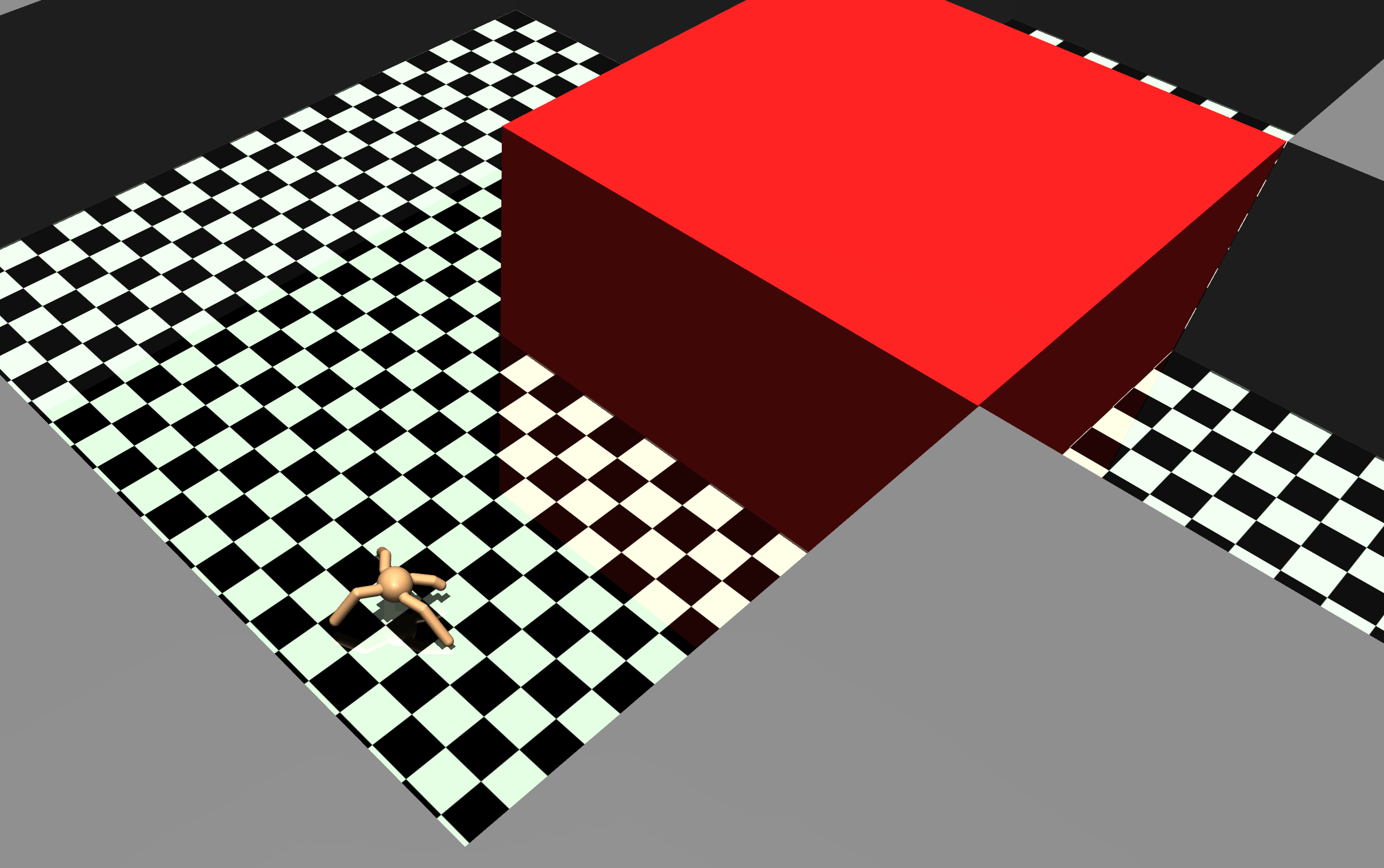}}} %
    
    \caption{MuJoCo Ant Complex Gait transfer learning task environments. We perform these transfer learning tasks with the Big Ant and the Amputated Ant.}%
        \label{fig:complex_ens}%

    \end{center}

    \vskip -0.2in

\end{figure}

We also utilize an Ant-Maze environment \cite{pmlr-v78-florensa17a} where we have a simple maze with a goal at the end.  The agent receives a reward of $+1$ if it reaches the goal and $0$ elsewhere. The ant must learn to make two turns in the maze, the first is down the hallway for one step and then a turn towards the goal.  Again, we see hierarchical behavior in this task: we can think of sub-goals consisting of \emph{learning to exit the first hall of the maze}, then \emph{making the turn} and finally \emph{going down the final hall towards the goal}.  The two complex environments are shown in Figure \ref{fig:complex_ens}.

Table \ref{complexResults} shows that oIRL performs better than AIRL in all of the complex hierarchical transfer tasks.  In some tasks such as the Maze environment, AIRL fails to have any or very few successful runs while our method achieves reasonably high reward.  In the BigAnt push task, AIRL achieves only very minimal reward where oIRL succeeds to perform the task in some cases.

\begin{table*}[h]
\caption{The mean reward obtained (higher is better) over 100 runs for the MuJoCo Ant Complex Gait transfer learning tasks. We also show the results of PPO optimizing the ground truth reward.}
\label{complexResults}
\vskip 0.15in
\begin{center}
\begin{small}
\begin{sc}
\begin{tabular}{lccccr}
\toprule
    & Big Ant Maze & Amputated Ant Maze & Big Ant Push & Amputated Ant Push \\
\midrule
    AIRL (Primitive) & 0.28  & 0.14   & 0.02  & 0.17    \\
    2 Options oIRL    & \textbf{0.62} & 0.29     & 0.46   & 0.34   \\
    4 Options oIRL    & 0.55     & \textbf{0.31}  & \textbf{0.55}  & \textbf{0.41} \\
    \textbf{Ground Truth} & 0.96  & 0.98 & 0.90  & 0.86
\end{tabular}
\end{sc}
\end{small}
\end{center}
\vskip -0.1in
\end{table*}

\begin{figure*}[]%
    \begin{center}
    \subfloat[Ant]{\includegraphics[width=4.5cm]{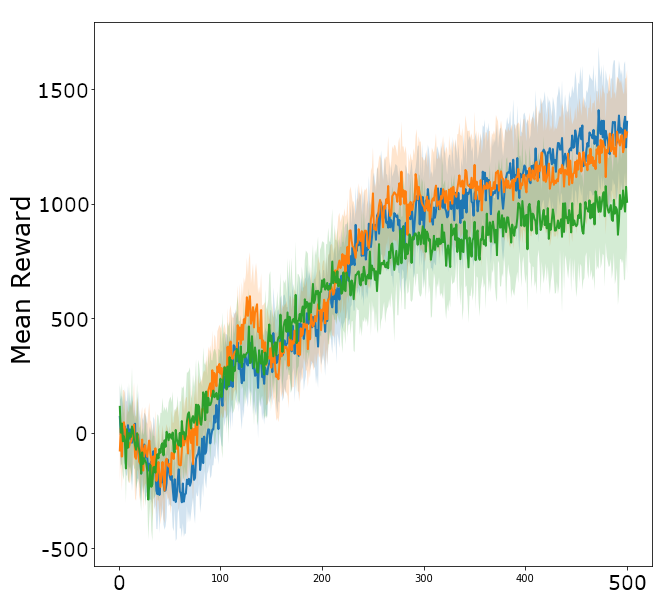}} \hspace{0.5cm}%
    \subfloat[Half Cheetah]{{\includegraphics[width=3.9cm]{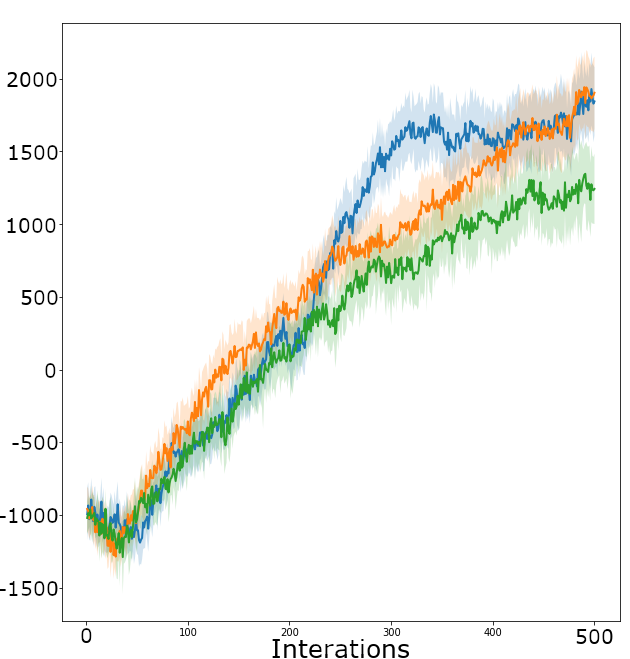}}}\hspace{0.5cm}%
        \subfloat[Walker]{{\includegraphics[width=4.21cm]{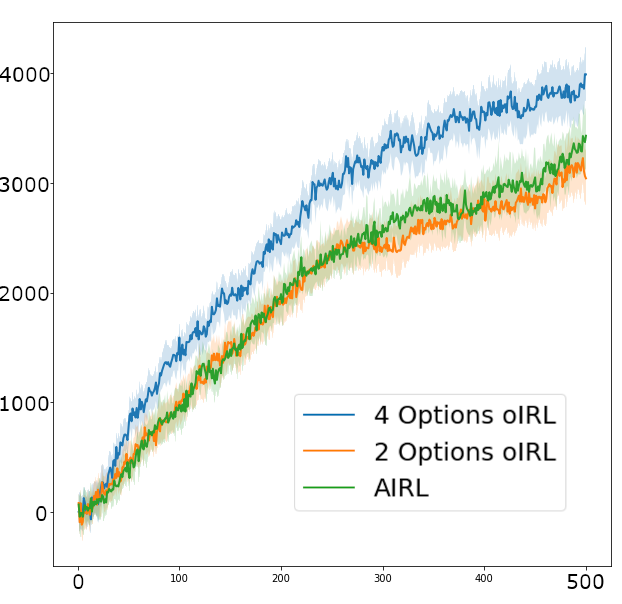}}}%

    \caption{MuJoCo Continuous control locomotion tasks showing the mean reward (higher is better) achieved over 500 iterations of the benchmark algorithms for 10 random seeds.  The shaded area represents the standard deviation.}%
            \label{fig:loco}%

    \end{center}

    \vskip -0.2in
\end{figure*}

\subsection{MuJoCo Continuous Control Benchmarks}
We also test our algorithm on a number of robotic continuous control benchmark tasks.  These tasks do not involve transfer.

We show the  plots of the average reward for each iteration during training in Figure \ref{fig:loco}.  Achieving a higher reward in fewer iterations is better for these experiments.  We examine the Ant, the Half Cheetah, the and Walker MuJoCo gait/locomotion tasks.  We run these experiments with 10 random seeds.  The results are quite similar between the benchmarks.  Using a policy over options shows reasonable improvements in each task.

\section{Discussion}
This work presents Option-Inverse Reinforcement Learning (oIRL), the first hierarchical IRL algorithm with disentangled rewards.  We validate oIRL on a wide variety of tasks, including transfer learning tasks, locomotion tasks, complex hierarchical transfer RL environments and GridWorld transfer navigation tasks and compare our results with the state-of-the-art algorithm. Combining options with a disentangled IRL framework results in highly portable policies.  Our empirical studies show clear and significant improvements for transfer learning.  The algorithm is also shown to perform well in continuous control benchmark tasks.

For future work, we wish to test other sampling methods (e.g., Markov-chain Monte Carlo) to estimate the implicit discriminator-generator pair's distribution in our GAN, such as Metropolis-Hastings GAN \cite{pmlr-v97-turner19a}.  We also wish to investigate methods to reduce the computational complexity for the step of computing the recursive loss function, which requires simulating some short trajectories, lowering the variance.  Analyzing our algorithm using physical robotic tests for tasks that require multiple sub-tasks would be an interesting future course of research. 


\nocite{langley00}

\bibliography{example_paper}
\bibliographystyle{icml2020}


\onecolumn
\appendix
\section{Option Transition Probabilities}
It is useful to redefine transition probabilities in terms of options.  Since at each step we have a additional consideration, we can continue following the policy of the current option we are in or terminate the option with some probability, sample a new option and follow that option's policy from a stochastic policy dependent on states.  We have
\begin{equation}
\begin{split}
P(s_{t+1}, \omega_{t+1} | s_t,\omega_t) &= \sum_{a \in \mathcal{A}} \pi_{\omega,\alpha}(a | s_t) P(s_{t+1} | s_t,a)((1-\beta_{\omega_t,\delta}(s_{t+1}))\textbf{1}_{\omega_t=\omega_{t+1}} + \\
&\beta_{\omega_t,\delta}(s_{t+1})\pi_{\Omega,\zeta}(\omega_{t+1} | s_{t+1}))
\end{split}
\end{equation}
\begin{equation}
\begin{split}
 P(s_{t+1},\omega_{t+1} | s_t) & =\sum_{\omega \in \Omega} \pi_{\zeta,\Omega}(\omega|s_t)\sum_{a \in \mathcal{A}} \pi_{\omega,\alpha}(a | s_t) P(s_{t+1} | s_t,a)((1-\beta_{\omega_t,\delta}(s_{t+1}))\textbf{1}_{\omega_t=\omega_{t+1}} + \\
&\beta_{\omega_t,\delta}(s_{t+1})\pi_{\Omega,\zeta}(\omega_{t+1} | s_{t+1}))
\end{split}
\end{equation}
\begin{equation}
\begin{split}
P(s_{t+1}, w_{t+1} | s_t,\omega_t,a_t) &=P(s_{t+1} | s_t,a_t)((1-\beta_{\omega_t,\delta}(s_{t+1}))\textbf{1}_{\omega_t=\omega_{t+1}} + \\
& \beta_{\omega_t,\delta}(s_{t+1})\pi_{\Omega,\zeta}(\omega_{t+1}|s_{t+1}))
\end{split}
\end{equation} 

\section{MLE Objective for IRL Over Options}
\label{sec:result}

We can define a discounted return recursively for a policy over options in a similar manner to the transition probabilities. Consider the policy over options based on the probabilities of terminating or continuing the option policies given a reward approximator $\hat{r}_{\theta}(s,a)$ for the state-action reward.

\begin{equation}
\begin{aligned}
& R^{\Omega}_{\zeta,\theta,\alpha,\delta}(s) = \EX_{\omega \in \Omega}[R^{}_{\theta,\alpha,\delta}(s,\omega)] \\
& R^{}_{\theta,\alpha,\delta}(s,\omega) = \EX_{a \in A}[R^{}_{\theta,\delta}(s,\omega,a)] \\
& R^{\Omega}_{\zeta,\theta,\delta}(s,a) = \EX_{\omega \in \Omega}[R^{}_{\theta,\delta}(s,\omega,a)] \\ 
& R^{}_{\theta,\delta}(s,\omega,a) = \EX \biggl[\hat{r}_{\omega,\theta}(s,a) +\\
& \gamma \sum_{s^{'} \in S}  P(s^{'}|s,a,\omega)(\beta_{\omega,\delta}(s^{'})R^{\Omega}_{\zeta,\theta,\alpha,\delta}(s^{'})+(1-\beta_{\omega,\delta}(s^{'}))R^{}_{\theta,\alpha,\delta}(s^{'},\omega))\biggr],
\end{aligned}
\end{equation}

These formulations of the reward function account for option transition probabilities, including the probability of terminating the current option and therefore selecting a new one according to the policy over options.

With $\omega_0$ selected according to $\pi_{\zeta,\Omega}(\omega | s)$, we can define a parameterization of the discounted return $R$ in the style of a maximum causal entropy RL problem with objective $\max_{\theta} \EX_{\tau \sim \mathcal{D}}[\log(p_{\theta}(\tau))]$, where

\begin{equation}
p_{\theta}(\tau) \sim p (s_0,\omega_0)\prod_{t=0}^{T-1} P(s_{t+1}, \omega_{t+1} | s_t, \omega_t, a_t)e^{R^{}_{\theta,\delta}(s_t,\omega_t,a_t)}.
\end{equation}

\textbf{MLE Derivative}

We can write out our MLE objective for our generator. We may or may not know the option trajectories in our expert demonstrations, but they are estimated below according to the policy over options. This is defined similarly in \cite{airl} and \cite{gcl} as 
$J(\theta) = \EX_{\tau \sim \tau^E}[\sum_{t=0}^{T} R^{\Omega}_{\zeta,\theta,\delta}(s_t,a_t)] - \EX_{p_{\theta}}[\sum_{t=0}^{T} \sum_{\omega \in \Omega}\pi_{\zeta,\Omega}(\omega | s_t) R^{}_{\theta,\delta}(s_t,\omega,a_t)]$.  The full derivation is shown as (with generator $p_\theta$):

\begin{equation}
\begin{aligned}
&J(\theta)=    \EX_{\tau \sim \tau^E}\left[\log(p_\theta(\tau))\right] \\
& = \EX_{\tau \sim \tau^E}\left[R^{}_{\theta,\delta}(s_t,\omega,a_t)\right] -  \log(Z_\theta)  \\
& = \EX_{\tau \sim \tau^E}\left[\sum_{t=0}^{T} \sum_{\omega \in \Omega}\pi_{\zeta,\Omega}(\omega | s_t) R^{}_{\theta}(s_t,\omega,a_t)\right] -  \log(Z_\theta)  \\
& \approx \EX_{\tau \sim \tau^E}\left[\sum_{t=0}^{T}  \sum_{\omega \in \Omega}\pi_{\zeta,\Omega}(\omega | s_t) R^{}_{\theta,\delta}(s_t,\omega,a_t)\right] - \EX_{p_{\theta}}\left[\sum_{t=0}^{T}  \sum_{\omega \in \Omega}\pi_{\zeta,\Omega}(\omega | s_t) R^{}_{\theta,\delta}(s_t,\omega,a_t)\right] \\
& = \EX_{\tau \sim \tau^E}\left[\sum_{t=0}^{T} R^{\Omega}_{\zeta,\theta,\delta}(s_t,a_t)\right] - \EX_{p_{\theta}}\left[\sum_{t=0}^{T} R^{\Omega}_{\zeta,\theta,\delta}(s_t,a_t)\right]
\end{aligned}
\end{equation}

We go from Line 4 to 5 seeing $R^{\Omega}_{\zeta,\theta,\delta}(s_t,a_t)=\sum_{\omega \in \Omega}\pi_{\zeta,\Omega}(\omega | s_t) R^{}_{\theta,\delta}(s_t,\omega,a_t)$.

Now, we take the gradient of the MLE objective w.r.t $\theta$ yields,

\begin{equation}
\begin{aligned}
&\frac{\partial}{\partial \theta}J(\theta)=  \EX_{\tau \sim \tau^E}\biggl[\frac{\partial}{\partial \theta}\log(p_\theta(\tau))\biggr] \\
& \frac{\partial}{\partial \theta}J(\theta)= \EX_{\tau \sim \tau^E}\biggl[\sum_{t=0}^{T}\sum_{\omega \in \Omega}\pi_{\zeta,\Omega}(\omega | s_t)\frac{\partial}{\partial \theta}R^{}_{\theta,\delta}(s_t,\omega,a_t)\biggr] - \frac{\partial}{\partial \theta} \log(Z_\theta)  \\
& \approx \EX_{\tau \sim \tau^E}\biggl[\sum_{t=0}^{T}  \frac{\partial}{\partial \theta}R^{\Omega}_{\zeta,\theta,\delta}(s_t,a_t)\biggr] - \EX_{p_{\theta}}\biggl[\sum_{t=0}^{T}  \frac{\partial}{\partial \theta} R^{\Omega}_{\zeta,\theta,\delta}(s_t,a_t)\biggr]
\end{aligned}
\end{equation}

Remark we define $p_{\theta,t}(s_t,a_t)=\int_{s_{t^{'} \neq t}, a_{t^{'} \neq t}} p_{\theta}(\tau) ds_{t'} da_{t'}$ as the state action marginal at time $t$.

\begin{equation}
\frac{\partial}{\partial \theta}J(\theta) = \sum_{t=0}^{T} \EX_{\tau \sim \tau^E}\biggl[\frac{\partial}{\partial \theta}R^{\Omega}_{\zeta,\theta,\delta}(s_t,a_t)\biggr] - \EX_{p_{\theta,t}}\biggl[\frac{\partial}{\partial \theta}R^{\Omega}_{\zeta,\theta,\delta}(s_t,a_t)\biggr]
\end{equation}

We perform importance sampling over the hard to estimate generator density.  We make an importance sampling distribution $\mu_{t,w}(\tau)$ for option $w$.  

We sample a mixture policy $\mu_{\omega}(a|s)$ defined as $\frac{1}{2} \pi_{\omega}(a|s) + \frac{1}{2}\hat{p}_{\omega}(a | s)$ and $\hat{p}_{\omega}(a|s)$ is a rough density estimate trained on the demonstrations.  We wish to minimize the $D_{KL}(\pi_w(\tau) | p_{\omega}(\tau))$.  KL refers to the Kullback–Leibler divergence metric between two probability distributions. Our new gradient is:

\begin{equation}
\frac{\partial}{\partial \theta}J(\theta) = \sum_{t=0}^{T} \EX_{\tau \sim \tau^E}[\frac{\partial}{\partial \theta}R^{\Omega}_{\zeta,\theta,\delta}(s_t,a_t)] - \EX_{\mu_{t}}\biggl[\sum_{\omega \in \Omega}\pi_{\zeta,\Omega}(\omega | s_t)\frac{p_{\theta,t,\omega}(s_t,a_t)}{\mu_{t,w}(s_t,a_t)}
\frac{\partial}{\partial \theta}R^{}_{\theta,\delta}(s_t,\omega,a_t)\biggr].
\end{equation}

Taking the derivative of the discounted option return results in

\begin{equation}
\begin{aligned}
& \frac{\partial}{\partial \theta}R^{\Omega}_{\zeta,\theta,\alpha,\delta}(s) = \EX_{}\biggl[\sum_{\omega \in \Omega}\pi_{\Omega,\zeta}(\omega|s)[\sum_{a \in A}[\pi_{w,\alpha}(a|s)\biggl(\frac{\partial}{\partial \theta}\hat{r}_{\omega,\theta}(s,a)
\\ + \gamma &\sum_{s^{'} \in S} P(s^{'}|s,a)(\beta_{\omega,\delta}(s^{'})\frac{\partial}{\partial \theta}R^{\Omega}_{\zeta,\theta,\alpha,\delta}(s^{'})+(1-\beta_{\omega,\delta}(s^{'}))\frac{\partial}{\partial \theta}R^{}_{\theta,\alpha,\delta}(s^{'},\omega))\biggr)]\biggr].
\end{aligned}
\end{equation}

\begin{equation}
\begin{aligned}
& \frac{\partial}{\partial \theta}R^{\Omega}_{\zeta,\theta,\delta}(s,a) = \EX_{}\biggl[\sum_{\omega \in \Omega}\pi_{\Omega,\zeta}(\omega|s)\biggl(\frac{\partial}{\partial \theta}\hat{r}_{\omega,\theta}(s,a)
\\ + \gamma &\sum_{s^{'} \in S} P(s^{'}|s,a)(\beta_{\omega,\delta}(s^{'})\frac{\partial}{\partial \theta}R^{\Omega}_{\zeta,\theta,\alpha,\delta}(s^{'})+(1-\beta_{\omega,\delta}(s^{'}))\frac{\partial}{\partial \theta}R^{}_{\theta,\alpha,\delta}(s^{'},\omega))\biggr)\biggr]
\end{aligned}
\end{equation}

\section{Discriminator Objective}

We formulate the discriminator as the odds ratio between the policy and the exponentiated reward distribution for option $\omega$ as in AIRL parameterized by $\theta$.  We have a discriminator for each option $\omega$ and generator option policy $\pi_w$,

\begin{equation}
D_{\theta,\omega} (s,a) = \frac{\exp(f_{\theta,\omega}(s,a))}{\exp(f_{\theta,\omega}(s,a)) + \pi_w(a | s)}.
\end{equation}

\subsection{Recursive Loss Formulation}

We minimize the cross-entropy loss between expert demonstrations and generated examples assuming we have the same number of options in the generated and expert trajectories.
We define the loss function $L_\theta$ as follows: 

\begin{equation}
\begin{aligned}
L_{\theta}(s,a,\omega) = -\EX_{\mathcal{D}}[\log(D_{\theta,\omega}(s,a))]-\EX_{\pi_{\Theta,t}}[\log(1-D_{\theta,\omega}(s,a))].
\end{aligned}
\end{equation}

The total loss for the entire trajectory can be expressed recursively as follows by taking expectations over the next options or states:
\begin{equation}\label{eq:L}
\begin{aligned} 
&L^{}_{\theta,\delta}(s,a,\omega) = l_{\theta}(s,a,\omega) + \gamma \sum_{s^{'} \in S}  P(s^{'}|s,a)(\beta_{w,\delta}(s^{'})L^{\Omega}_{\zeta,\theta,\alpha,\delta}(s^{'})+(1-\beta_{w,\delta}(s^{'}))L^{}_{\theta,\alpha,\delta}(s^{'},w)) \\  
& L^{}_{\theta,\alpha,\delta}(s,w) = \EX_{a \in A}[[L^{}_{\theta,\delta}(s,w,a)] \\ 
& L^{\Omega}_{\zeta,\theta,\delta}(s,a) = \EX_{w \in \Omega}[L^{}_{\theta,\delta}(s,w,a)] \\
&L^{\Omega}_{\zeta,\theta,\alpha,\delta}(s) = \EX_{\omega \in \Omega}[L_{\theta,\alpha,\delta}(s,\omega)] \\ 
\end{aligned}
\end{equation}
The agent wishes to minimize $L^{}_{\theta,\delta}$ to find its optimal policy.

We can let cost function $f_{\theta,w}(s,a)=L_{\theta,\delta}(s,\omega,a)$ as shown in AIRL and we have:

\begin{equation}
D_{\theta,\omega} = \frac{\exp(L_{\theta,\delta}(s,\omega,a))}{\exp(L_{\theta,\delta}(s,\omega,a)) + \pi_{\omega}(a | s)}
\end{equation}


\subsection{Optimization Criteria}


For a given option $\omega$, we can write the reward function $\hat{R^{}}_{\theta,\delta}(s,\omega,a)$ to be maximised,  as follows.  Note that $\theta$ parameterizes the state-action reward function estimate for option $\omega$. $-L^{D}$ is the negative discriminator loss.  We therefore turn our minimization problem into a maximization problem.  We define our objective similar to the GAN objective from AIRL:

\begin{equation}
\begin{aligned}
-L^{D}=\hat{R^{}}_{\theta,\delta}(s,\omega,a) = \log \left( D_{\theta,\omega}(s,a)\right) - \log \left(1- D_{\theta,\omega}(s,a)\right)
\end{aligned}
\end{equation}

Now we can write out our reward function in terms of the optimal discriminator

\begin{equation}
\begin{aligned}
& \hat{R^{}}_{\theta,\delta}(s,\omega,a) =  \log \left(\frac{\exp \left(-L_{\theta,\delta}(s,\omega,a)\right)}{\exp \left(-L_{\theta,\delta}(s,\omega,a)\right) + \pi_{\omega}(a | s)}\right) - \log\left(\frac{\pi_{\omega}(a | s)}{\exp \left(-L_{\theta,\delta}(s,\omega,a)\right) + \pi_{\omega}(a | s)}\right) \\
& = -L_{\theta,\delta}(s,\omega,a) - \log(\pi_{\omega}(a | s))
\end{aligned}
\end{equation}

The derivative of this reward function can now be computed as follows:

\begin{equation}
\begin{aligned}
&\frac{\partial}{\partial \theta} \hat{R^{}}_{\theta,\delta}(s,\omega,a) \approx \frac{\partial}{\partial \theta} -L_{\theta,\delta}(s,\omega,a)\\
& = \EX \left[\frac{\partial}{\partial \theta}r_{\omega,\theta}(s,a) + \gamma \sum_{s^{'} \in S}  P(s^{'}|s,a) \left(\beta_{\omega,\delta}(s^{'})\frac{\partial}{\partial \theta}-L^{\Omega}_{\zeta,\theta,\alpha,\delta}(s^{'})+(1-\beta_{\omega,\delta}(s^{'}))\frac{\partial}{\partial \theta}-L^{}_{\theta,\alpha,\delta}(s^{'},\omega)\right)\right] \\
& = \EX \left[\frac{\partial}{\partial \theta}r_{\omega,\theta}(s,a)\right] + \EX \biggr[\gamma \sum_{s^{'} \in S}  P(s^{'}|s,a)\biggl(\beta_{\omega,\delta}(s^{'})\frac{\partial}{\partial \theta}-L^{\Omega}_{\zeta,\theta,\alpha,\delta}(s^{'})+ \\
& (1-\beta_{\omega,\delta}(s^{'}))\frac{\partial}{\partial \theta}-L^{}_{\theta,\alpha,\delta}(s^{'},\omega) \biggr) \biggr] \\
\end{aligned}
\end{equation}

Writing out our discriminator objective yields:

\begin{equation}
\begin{aligned}
& -L^D = \sum_{t=0}^{T} \EX_{\tau \sim \tau^E}\bigg(\Big[\sum_{\omega \in \Omega}\pi_{\Omega,\zeta}(\omega|s_t)\log( D_{\theta,\omega}(s_t,a_t))\Big] + \\
&\EX_{\pi_t}\Big[\sum_{\omega \in \Omega}\pi_{\Omega,\zeta}(\omega|s_t)\log(1- D_{\theta,\omega}(s_t,a_t))\Big]\bigg) \\
& = \sum_{t=0}^{T} \EX_{\tau \sim \tau^E}\left[\sum_{\omega \in \Omega}\pi_{\Omega,\zeta}(\omega|s_t)\log \left( \frac{\exp(-L_{\theta,\delta}(s_t,\omega,a_t))}{\exp(-L_{\theta,\delta}(s_t,\omega,a_t)) + \pi_{\omega}(a_t | s_t)}\right)\right]\\
& + \EX_{\pi_t}\left[\sum_{\omega \in \Omega}\pi_{\Omega,\zeta}(\omega|s_t)\log \left(\frac{\pi_{\omega}(a_t | s_t)}{\exp(-L_{\theta,\delta}(s_t,\omega,a_t)) + \pi_{\omega}(a_t | s_t)}\right)\right]  \\
& = \sum_{t=0}^{T} \EX_{\tau \sim \tau^E}\left[\sum_{\omega \in \Omega}\pi_{\Omega,\zeta}(\omega|s_t)-L_{\theta,\delta}(s_t,\omega,a_t)\right] - \\
&\EX_{\tau \sim \tau^E}\left[\sum_{\omega \in \Omega}\pi_{\Omega,\zeta}(w|s_t)\log(\exp(-L_{\theta,\delta}(s_t,w,a_t)) + \pi_{\omega}(a | s_t))\right]\\
& + \EX_{\pi_t}\left[\sum_{\omega \in \Omega}\pi_{\Omega,\zeta}(\omega|s_t)\log(\pi_{\omega}(a_t | s_t))\right] - \\
&\EX_{\pi_t}\left[\sum_{\omega \in \Omega}\pi_{\Omega,\zeta}(\omega|s_t)\log(\exp(-L_{\theta,\delta}(s_t,\omega,a_t)) + \pi_{\omega}(a_t | s_t))\right]  \\
\end{aligned}
\end{equation}

We set a mixture of experts and novice as $\bar{\mu}$ observations.
\begin{equation}
\begin{aligned}
& = \sum_{t=0}^{T} \EX_{\tau \sim \tau^E}\left[\sum_{\omega \in \Omega}\pi_{\Omega,\zeta}(\omega|s_t)-L_{\theta,\delta}(s_t,\omega,a_t)\right] + \EX_{\pi_t}\left[\sum_{\omega \in \Omega}\pi_{\Omega,\zeta}(\omega|s_t)\log(\pi_{\omega}(a_t | s_t))\right] \\
& - 2\EX_{\bar{\mu}_t}\left[\sum_{\omega \in \Omega}\pi_{\Omega,\zeta}(\omega|s_t)\log \left(\exp(-L_{\theta,\delta}(s_t,\omega,a_t)) + \pi_{\omega}(a_t | s_t)\right)\right]  \\
\end{aligned}
\end{equation}

We can take the derivative w.r.t $\theta$ (state-action reward function estimate parameter):
\begin{equation}
\begin{aligned}
& \frac{\partial}{\partial \theta} (-L^D) = \sum_{t=0}^{T} \EX_{\tau \sim \tau^E}\left[\sum_{\omega \in \Omega}\pi_{\Omega,\zeta}(\omega|s_t)\frac{\partial}{\partial \theta}-L_{\theta,\delta}(s_t,\omega,a_t)\right]\\ & - \EX_{\bar{\mu}_t}\left[\sum_{\omega \in \Omega}\pi_{\Omega,\zeta}(\omega|s_t) \left(\frac{\exp(-L_{\theta,\delta}(s_t,\omega,a_t))}{\frac{1}{2}\exp(-L_{\theta,\delta}(s_t,\omega,a_t)) +\frac{1}{2} \pi_{\omega}(a_t | s_t)}\right)\frac{\partial}{\partial \theta}-L_{\theta,\delta}(s_t,\omega,a_t)\right]  \\
\end{aligned}
\end{equation}

We can multiply the top and bottom of the fraction in the mixture expectation by the state marginal $\pi_{\omega}(s_t) = \int_{a \in A} \pi_{\omega}(s_t,a_t)$.  This allows us to write  $\hat{p}_{\theta,t,\omega}(s_t,a_t) = \exp(L_{\theta,\delta}(s_t,\omega,a_t))\pi_{\omega,t}(s_t)$. Now we have an importance sampling.

\begin{equation}
\begin{aligned}
& \frac{\partial}{\partial \theta} (-L^D) = \sum_{t=0}^{T} \EX_{\tau \sim \tau^E}\left[\sum_{\omega \in \Omega}\pi_{\Omega,\zeta}(\omega|s_t)\frac{\partial}{\partial \theta}-L_{\theta,\delta}(s_t,\omega,a_t)\right]\\ & - \EX_{\bar{\mu}_t}\left[\sum_{\omega \in \Omega}\pi_{\Omega,\zeta}(\omega|s_t) \left(\frac{\hat{p}_{\theta,t,\omega}(s_t,a_t)}{\hat{\mu}_{t,\omega}(s_t,a_t)}\right)\frac{\partial}{\partial \theta}-L_{\theta,\delta}(s_t,\omega,a_t)\right]  \\
\end{aligned}
\end{equation}

It is now easy to see we have the same form as our MLE objective loss function, our loss (the function we approximate with the GAN) is the discounted reward for a state action pair with the expectation over options. We change the loss functions to reward functions to show this, as they are defined equivalently. 

\begin{equation}
\begin{aligned}
& \frac{\partial}{\partial \theta} (-L^D) = \sum_{t=0}^{T} \EX_{\tau \sim \tau^E}\left[\frac{\partial}{\partial \theta}R_{\zeta,\theta,\delta}(s_t,a_t)\right]\\ & - \EX_{\bar{\mu}_t}\left[\sum_{\omega \in \Omega}\pi_{\Omega,\zeta}(\omega|s) \left(\frac{\hat{p}_{\theta,t,\omega}(s_t,a_t)}{\hat{\mu}_{t,\omega}(s_t,a_t)}\right)\frac{\partial}{\partial \theta}R_{\theta,\delta}(s_t,\omega,a_t)\right] \\
\end{aligned}
\label{deriv-theta}
\end{equation}

In addition, we can decompose the reward into a state-action reward and a future discounted sum of rewards considering the policy over options as follows:

\begin{equation}
\begin{aligned}
& \frac{\partial}{\partial \theta} (-L^D) = \sum_{t=0}^{T} \underbrace{\EX_{\tau \sim \tau^E}\left [\sum_{\omega \in \Omega}\pi_{\Omega,\zeta}(\omega|s_t)\frac{\partial}{\partial \theta}r_{\omega,\theta}(s_t,a_t)\right]}_{\text{State-Action Reward}} \\
& + \EX_{\tau \sim \tau^E}\biggl[\sum_{\omega \in \Omega}\pi_{\Omega,\zeta}(\omega|s_t) \gamma\sum_{s_{t+1} \in S}  P(s_{t+1}|s_t,a_t)(\beta_{\omega,\delta}(s_{t+1})\frac{\partial}{\partial \theta}R^{\Omega}_{\zeta,\theta,\alpha,\delta}(s_{t+1})+ \\
&(1-\beta_{w,\delta}(s_{t+1}))\frac{\partial}{\partial \theta}R_{\theta,\alpha,\delta}(s_{t+1},\omega))\biggr] \\ 
& - \EX_{\bar{\mu}_t}\left[\sum_{\omega \in \Omega}\pi_{\Omega,\zeta}(\omega|s_t) \left(\frac{\hat{p}_{\theta,t,\omega}(s_t,a_t)}{\hat{\mu}_{t,\omega}(s_t,a_t)}\right)\frac{\partial}{\partial \theta}R_{\theta,\delta}(s_t,\omega,a_t)\right] \\
\end{aligned}
\end{equation}

\section{GAN Architecture}

\begin{figure}
    \centering
    \includegraphics[width=9.25cm]{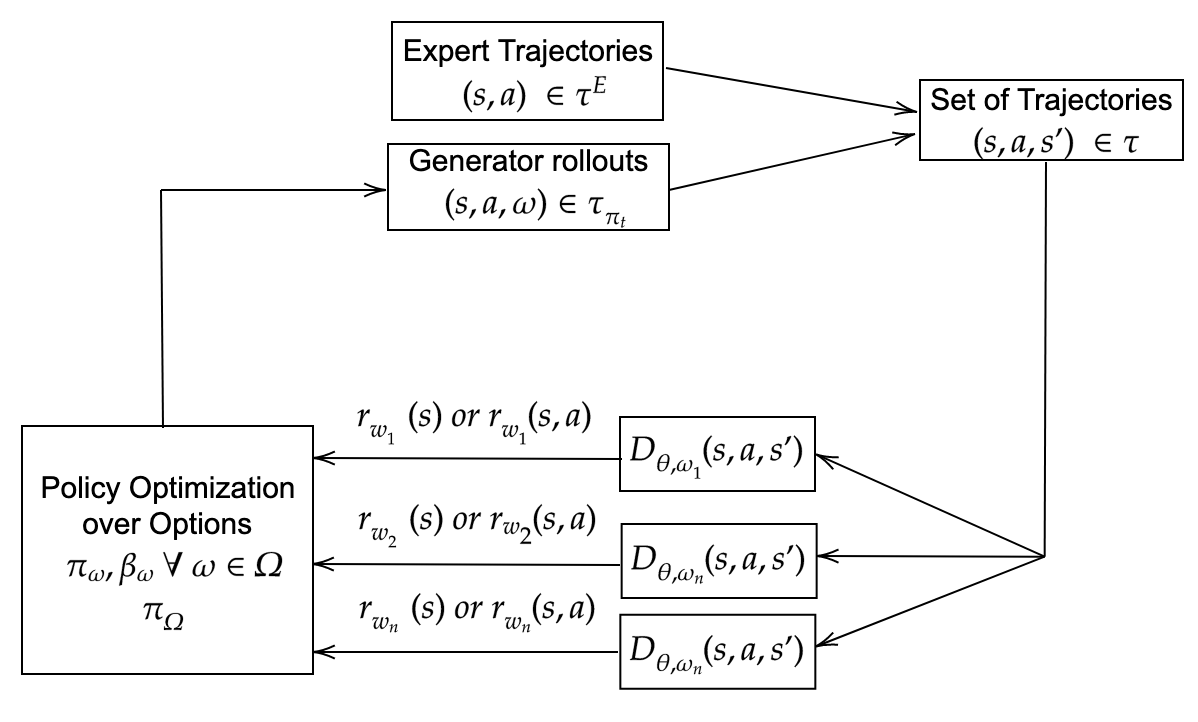}
    \caption{Architecture of GAN-IRL framework.}
    \label{fig:gan_arch}
\end{figure}

The architecture for our GAN-IRL framework is described in Figure \ref{fig:gan_arch}.

\section{Proof of Recoverable Rewards}

A substantial amount of this proof is derived from \cite{airl}.
    \vspace{3mm}

\textbf{Lemma 1:} $f_{\theta,\omega}(s,a)$ \textit{recovers the advantage.}

\textbf{Proof:} It is known that when $\pi_{\omega} = \pi_{\omega}^E$, we have achieved the global min of the discriminator objective.  The discriminator must then output 0.5 for all state action pairs.  This results in $\text{exp}(f_{\theta,\omega}(s,a))=\pi^E_\omega (a|s)$.  Equivalently we have $f^*_{\omega}(s,a)=\text{log }\pi_{\omega}^E(a|s)=A^*(s,a,\omega)$.
\vspace{3mm}

\textbf{Definition 1: Decomposability condition}. \textit{We first define 2 states $s_1, s_2$ as 1-step linked under dynamics $T(s'|s,a)$ if there exists a state $s$ that can reach $s_1$ and $s_2$ with non-zero probability in one timestep.  The transitivity property holds for the linked relationship.  We can say that if $s_1$ and $s_2$ and linked, $s_2$ and $s_3$ are linked then $s_1$ and $s_3$ must also be linked.}

\textit{The Decomposability condition for transition dynamics $T$ holds if all states in the MDP are linked with all other states.}

    \vspace{3mm}

\textbf{Lemma 2:} 
   \textit{For an MDP, where the decomposability condition holds for all dynamics.  For arbitrary functions $a(s),b(s),c(s),d(s)$, if for all $s$ and $s'$}
    
    \begin{equation}
        a(s) + b(s') = c(s) + d(s')
        \label{esd}
    \end{equation}
    
    \textit{and for all $s$}
    
    \begin{equation}
        a(s)=c(s) + \text{const}_s
    \end{equation}
    \begin{equation}
        b(s) = d(s) + \text{const}_s,
    \end{equation}
    
\textit{where $\text{const}_s$ is a constant dependent with respect to state  $s$.}

\textbf{Proof:} If we rearrange Equation \ref{esd}, we can obtain the quality $a(s)-c(s)=b(s')-d(s')$. 
    
    Now we define $f(s) = a(s) - c(s)$. Given our equality, we have $f(s) = a(s) - c(s)= b(s')-d(s')$. This holds for some function dependent on $s$.  
    
    To represent this, $b(s') - d(s')$ must be equal to a constant (with the constant's value dependent on the state $s$) for all one-step successor states $s'$ from $s$. 
    

    
    


    \vspace{3mm}

    Now, under decomposability, all one step successor states ($s'$) from $s$ must be equal through the transitivity property so $b(s')-d(s')$ must be a constant with respect to state $s$.  Therefore, we can write $a(s) = c(s) + \text{const}_+s$ for an arbitrary state $s$ and functions $b$ and $d$. 
    
    Substituting this into the Equation \ref{esd}, we can obtain $b(s)=d(s) + \text{const}_s$.  This completes our proof.

    \vspace{4mm}
    \textbf{Inductive proof for any successor state}\\
Let us consider for any MDP and any arbitrary functions $a(\cdot)$, $b(\cdot)$, $c(\cdot)$ and $d(\cdot)$,
\begin{equation}\label{eq:decomposibility}
a(s) + b(S^{(k)}) = c(s) + d(S^{(k)}),
\end{equation}
where $S^{(k)}$ is the $k$-th successor state reached in $k$ time-steps from the current state. Let us denote by $T^{\pi, (k)}(s, S^{(k)})$ the probability of transitioning from state $s$ to $S^{(k)}$ in $k$ steps using policy $\pi$. Then, we can express $T^{\pi, (k)}(s, S^{(k)})$ recursively as follows:
\begin{equation}\label{eq:T_pi_k}
    T^{\pi, (k)}(s, S^{(k)}) = \sum_{s' \in \mathcal S} T^{\pi, (k-1)}(s,  s')T^\pi(s', S^{(k)}),
\end{equation}
where $T^\pi(s', S^{(k)})$ is the one-step transition probability from state $s'$ to state $S^{(k)}$ (by definition of the Bellman operator).

Denote by $P(S^{(k)})$ the probability of landing in state $S^{(k)}$ in $k$ steps from any current state. We can write $P(S^{(k)})$ using~\eqref{eq:T_pi_k} as follows:
\begin{equation}\label{eq:P_s_k}
    P(S^{(k)}) \coloneqq \sum_{s \in \mathcal{S}} T^{\pi, (k)}(s, S^{(k)}) \mu(s),
\end{equation}
where $\mu$ is the state-distribution.

The unbiased estimator $\hat s^{(k)}$ of an unknown successor state $S^{(k)}$ is given by:
\begin{equation}\label{eq:hat_s_k}
    \hat s^{(k)} \coloneqq \mathbb{E} (S^{(k)}) = \sum_{s^{(k)} \in \mathcal S} s^{(k)} P(S^{(k)}), 
\end{equation}
where $P(S^{(k)})$ is given in~\eqref{eq:P_s_k}.

Now, replacing $S^{(k)}$ in~\eqref{eq:decomposibility} with its unbiased estimator $\hat s^{(k)}$ as given by~\eqref{eq:hat_s_k}, we have
\begin{equation}
    a(s) - c(s)  = b(\hat s^{(k)}) - d(\hat s^{(k)})
    \stackrel{(a)}{=} f(k), \label{eq:decomposibility_s_hat}
\end{equation}
for some function $f$, where $(a)$ holds since $\hat s^{(k)}$ depends only on $k$. Thus, we get $a(s) = c(s) + \text{const.}$ and $b(s) = d(s) + \text{const.}$ where the constant is with respect to the state $s$. 
    
    \vspace{3mm}
\textbf{Theorem 1:}
\textit{Suppose we have, for a MDP where the decomposability condition holds,}
\begin{equation}
    f_{\theta,\omega}(s,a,s')=g_\omega(s,a) + \gamma h_{\Phi}(s') - h_{\Phi}(s) 
\end{equation}
\textit{where $h_{\Phi}$ is a shaping term. If we obtain the optimal $f^*_{\theta,\omega}(s,a,s')$, with a reward approximator $g^*_{\omega}(s,a)$. Under deterministic dynamics the following holds}
\begin{equation}
    g^*_\omega(s,a) + \gamma h^*_{\Phi}(s') - h^*_{\Phi}(s) = r^*_\omega(s)+\gamma V^*_\Omega(s') - V^*_{\Omega}(s)
\end{equation}

and

\begin{equation}
    g^*_{\omega}(s) = r^*_\omega(s) + c_\omega.
\end{equation}

\textbf{Proof:} We know $f^*_\omega(s,a,s')=  A^*(s,a,\omega) = Q^*(s,a,\omega) - V_{\Omega}^*(s) = r^*_\omega(s)+\gamma V^*_{\Omega}(s') - V^*_{\Omega}(s)$.  We can substitute the definition of $f^*_\omega(s,a,s')$ to obtain our Theorem.

Where $Q(s,\omega)=\sum_{a \in \mathcal{A}}\pi_{\omega,\alpha}(a|s)[r_{\omega,\theta}(s,a)+\gamma \sum_{s' \in \mathcal{S}}P(s'|s,a) \left ((1-\beta_{\delta,\omega}(s'))Q(s',\omega) + \beta_{\delta,\omega}(s')V_{\Omega}(s') \right)]$ and $V_{\Omega}(s)=\sum_{\omega \in \Omega}\pi_{\Omega,\zeta}(\omega|s)Q(s,\omega)$

$Q(s,a,\omega) = \pi_{\omega,\alpha}(a|s)[r_{\omega,\theta}(s,a)+\gamma \sum_{s' \in \mathcal{S}}P(s'|s,a) \left ((1-\beta_{\delta,\omega}(s'))Q(s',\omega) + \beta_{\delta,\omega}(s')V_{\Omega}(s') \right)]$

which holds for all $s$ and $s'$.  Now we apply Lemma 2.  We say that $a(s) = g^*_{\omega}(s) - h^*_{\Phi}(s), b(s')= \gamma h^*_{\Phi}(s'), c(s) = r(s) - V^*_{\Omega}(s)$ and $d(s') = \gamma V^*_\Omega(s')$ and rearrange according to Lemma 2.  We therefore have our results that $g^*_{\omega}(s) = r_\omega(s) + c_\omega$.  Where $c_\omega$ is a constant.

\section{Proof of Convergence}

\textbf{Definition 2: Reward Approximator Error.} \textit{From Theorem 1, we can see that our reward approximator $g^*_{\omega}(s) = r_\omega(s) + c_\omega$. We define a reward approximator error over all options as $\delta_r = \sum_{\omega \in \Omega}\pi_{\Omega}(\omega) |{g}^*_\omega(s)-r^*(s)|$. This error is bounded by}

\begin{equation}
\delta_r=\sum_{\omega \in \Omega}\pi_{\Omega}(\omega) |g^*_{\omega}(s)-r^*(s)| \leq  \max_{\omega \in \Omega} c_\omega
\end{equation}

\textit{By definition of $g^*_{\omega}(s)$.}
\vspace{3mm}

\textbf{Lemma 3: }\textit{The Bellman operator for options in the IRL problem is a contraction.}

\textbf{Proof:} We prove this by Cauchy-Schwarz and the definition of the sup-norm.  We must define this inequality in terms of the IRL problem where we have a reward estimator $\hat{g}_{{\theta}_{\omega}}(s)$ under our learned parameter $\theta$ and an optimal reward estimator $r^*(s)$.

\begin{equation}
\begin{split}
    & ||Q_{\pi_{\Omega,t}}(s,\omega)-Q^*(s,\omega)||_{\infty}\\
    & = || \hat{g}_{\theta}(s) + \gamma \sum_{s'\in \mathcal{S}}  P(s'|s,a) ((1-\beta(s')Q_{\pi_{\Omega,t}}(s',\omega)+\beta(s') \max_{\omega \in \Omega} Q_{\pi_{\Omega,t}}(s',\omega)) - \\
    & r^*(s) + \gamma \sum_{s'\in \mathcal{S}}  P(s'|s,a) ((1-\beta(s')Q^*(s',\omega)+\beta(s') \max_{\omega \in \Omega}Q^*(s',\omega))||_{\infty}\\
    &=|| \hat{g}_{\theta}(s) - r^*(s) + \sum_{s' \in \mathcal{S}} P(s'|s,a)[(1-\beta(s'))(Q_{\pi_{\Omega,t}}(s',\omega)-Q^*(s',\omega))] + \\
    & [\beta(s')(\max_{\omega \in \Omega} Q_{\pi_{\Omega,t}}(s',\omega)-\max_{\omega \in \Omega}Q^*(s',\omega))]||_{\infty} \\
    & =  || \sum_{s' \in \mathcal{S}} P(s'|s,a)[(1-\beta(s'))(Q_{\pi_{\Omega,t}}(s',\omega)-Q^*(s',\omega))] + \\
    & [\beta(s')(\max_{\omega \in \Omega} Q_{\pi_{\Omega,t}}(s',\omega)-\max_{\omega \in \Omega}Q^*(s',\omega))]||_{\infty} + \max_{\omega \in \Omega} c_\omega \\
    & \leq  \sum_{s' \in \mathcal{S}}P(s'|s,a) \max_{s'',\omega''}|| Q_{\pi_{\Omega,t}}(s'',\omega'')-Q^*(s'',\omega'')||_{\infty} +  \max_{\omega \in \Omega} c_\omega \\
    & \leq  \gamma \max_{s'',\omega''}|| Q_{\pi_{\Omega,t}}(s'',\omega'')-Q^*(s'',\omega'')||_{\infty} + \max_{\omega \in \Omega} c_\omega\\
\end{split}
\end{equation}
This is given by Lemma 3 and \cite{Sutton:1999} [Theorem 3].

Giving our results $ \max_{s'',\omega''}|Q_{\pi_{\Omega,t}}(s,\omega)-Q^*(s,\omega)| \leq \epsilon + \max_{\omega \in \Omega} c_\omega$. For $\epsilon \in \mathbb{R}_{>0}$
    \vspace{3mm}

\textbf{Theorem 2:} $g_{\theta}(s) + \gamma Q(s',\omega)$ \textit{converges to} $Q^*$.

\textbf{Proof:} We know $g_{\theta}(s) \rightarrow g^*_{\theta}(s) = r^*(s) + \text{const}$. Given this we can show by Cauchy-Schwarz:

\begin{align}
&|\mathbb{E}[g_{\theta}(s)] + \gamma \mathbb{E}[Q(s',\omega)|s] - Q^*(s',\omega)| \notag \\
& = |\mathbb{E}[g_{\theta}(s)] + \gamma \sum_{s' \in \mathcal{S}}P(s'|s,a)((1-\beta_\omega(s'))Q(s'\omega) +\beta_\omega(s')V_{\Omega}(s')) \notag \\
&-r^*(s)-\sum_{s'\in \mathcal{S}}P(s'|s,a)((1-\beta_{\omega}(s')Q^*(s',\omega))+\beta_{\omega}(s')\max_{\omega \in \Omega}Q^*(s',\omega)| \notag \\
&  = |\mathbb{E}[g_{\theta}(s)]-r^*(s) + \gamma \sum_{s' \in \mathcal{S}}P(s'|s,a)[ \beta_\omega(s')[ \max_{\omega\in\Omega}Q(s'\omega)-\max_{\omega \in \Omega}Q^*(s',\omega)] \notag \\
&+(1-\beta_\omega(s'))[Q(s'\omega)-Q^*(s',\omega)]]|\notag \\
& \stackrel{(a)}{\leq} (\max_{\omega \in \Omega} c_\omega)| \gamma \sum_{s' \in \mathcal{S}}P(s'|s,a)[ \max_{s'',\omega''} ||Q(s'',\omega'')-Q^*(s'',\omega'')|]| \notag \\
& \stackrel{(b)}{\leq} (\max_{\omega \in \Omega} c_\omega) (\epsilon +\max_{\omega \in \Omega} c_\omega) \gamma \sum_{s' \in \mathcal{S}}P(s'|s,a) \notag \\
& \leq (\max_{\omega \in \Omega} c_\omega)(\epsilon+\max_{\omega \in \Omega} c_\omega) \gamma, \label{eq:contraction}
\end{align}
where $(a)$ follows from Lemma 3 and $(b)$ holds since $\sum_{s' \in \mathcal{S}}P(s'|s,a) \leq 1$. 

\section{Parameters for Experiments}
\label{sece}

\subsection{MuJoCo Tasks}

For these experiments, we use PPO to obtain an optimal policy given our ground truth rewards for 2 million iterations and 20 million on the complex tasks.  This is used to obtain the expert demonstrations.  We sample 50 expert trajectories. PPOC is used for the policy optimization step for the policy over options. We tune the deliberation cost hyper-parameter via cross-validation. The optimal deliberation cost found was $0.1$ for PPOC.  We also use state-only rewards for the policy transfer tasks.  The hyperparameters for our policy optimization are given in Table \ref{hyperopt}.

Our discriminator is a neural network with the optimal architecture of 2 linear layers of 50 hidden states, each with ReLU activation followed by a single node linear layer for output.  We also tried a variety of hidden states including 100 and 25 and tanh activation during our hyperparameter optimization step using cross-validation.

The policy network has 2 layers of 64 hidden states.  A batch size of 64 or 32 is used for 1 and any number of options greater than 1 respectively.  No mini-batches are used in the discriminator since the recursive loss must be computed.  There are 2048 timesteps per batch.  Generalized Advantage Estimation is used to compute advantage estimates.   We list additional network parameters in the next section.  The output of the policy network gives the Gaussian mean and the standard deviation. This is the same procedure as in \cite{schulman2017proximal}.

\begin{table}[tbh]
    \caption{Policy Optimization parameters for MuJoCo}
    \label{hyperopt}
  \begin{center}
    \begin{tabular}{lr}
      \toprule
      Parameter & Value  \\
      \midrule
      Discr. Adam optimizer learning rate                                   & $1\cdot 10^{-3}$      \\
    Adam $\epsilon$                                   & $1\cdot 10^{-5}$      \\
     PPOC Adam optimizer learning rate  & $3\cdot 10^{-4}$ \\
          GAE $\lambda$  & $0.95$ \\

      Entropy coefficient
      & $10^{-2}$\\
      value loss coefficient & 0.5\\
            discount                                        & 0.99      \\

      batch size for PPO &64 or 32\\
    PPO epochs &10\\

      entropy coefficient                    & $10^{-2}$      \\
      clip parameter &0.2
             \\
    
      \bottomrule
    \end{tabular}
  \end{center}
\end{table}

\subsection{MuJoCo Continuous Control Tasks}

In this section, we describe the structure of the objects that gait in the continuous control benchmarks and the reward functions.  For the transfer learning tasks, we use the same reward function described here for the Ant.

\textbf{Walker:} The walker is a planar biped. There are 7 rigid links comprised of legs, a torso.  This includes 6 actuated joints. This task is particularly prone to falling. The state space is of 21 dimensions.  The observations in the states include joint angles, joint velocities, the center of mass's coordinates. The reward function is $r(s,a) =v_x - 0.005 ||a||^2_2$. The termination condition occurs when $z_{\text{body}} < 0.8, z_{\text{body}} >2.0 $ or $||\theta_y||>1.0$.

\textbf{Half-Cheetah:} The half-cheetah is a planar biped also like the Walker. There are 9 rigid links comprised of 9 actuated joints, a leg and a torso. The state space is of 20 dimensions.  The observations include joint angles, the center of mass's coordinates, and joint velocities. The reward function is $r(s, a) = v_x - 0.005 ||a||_2^2$. There is no termination condition.

\textbf{Ant}: The ant has four legs with 13 rigid links in its structure. The legs have 8 actuated joints. The state space is of 125 dimensions. This includes joint angles, joint velocities, coordinates of the center of mass, the rotation matrix for the body, and a vector of contact forces. The function is $r(s, a) = v_x - 0.005 ||a||^2_2 - C_{\text{contact}} + 0.05$, where $C_{\text{contact}}$ is a penalty for contacts to the ground.  This is $5 \times 10^{-4}  ||F_{\text{contact}}||_2^2$. $F_{\text{contact}}$ is the contact force.  It's values are clipped to be between 0 and 1. The termination condition occurs $z_{\text{body}} < 0.2$ or $z_{\text{body}} > 1.0$.

\subsection{MiniGrid Tasks}
\begin{figure}
    \centering
    \includegraphics[width=5.25cm]{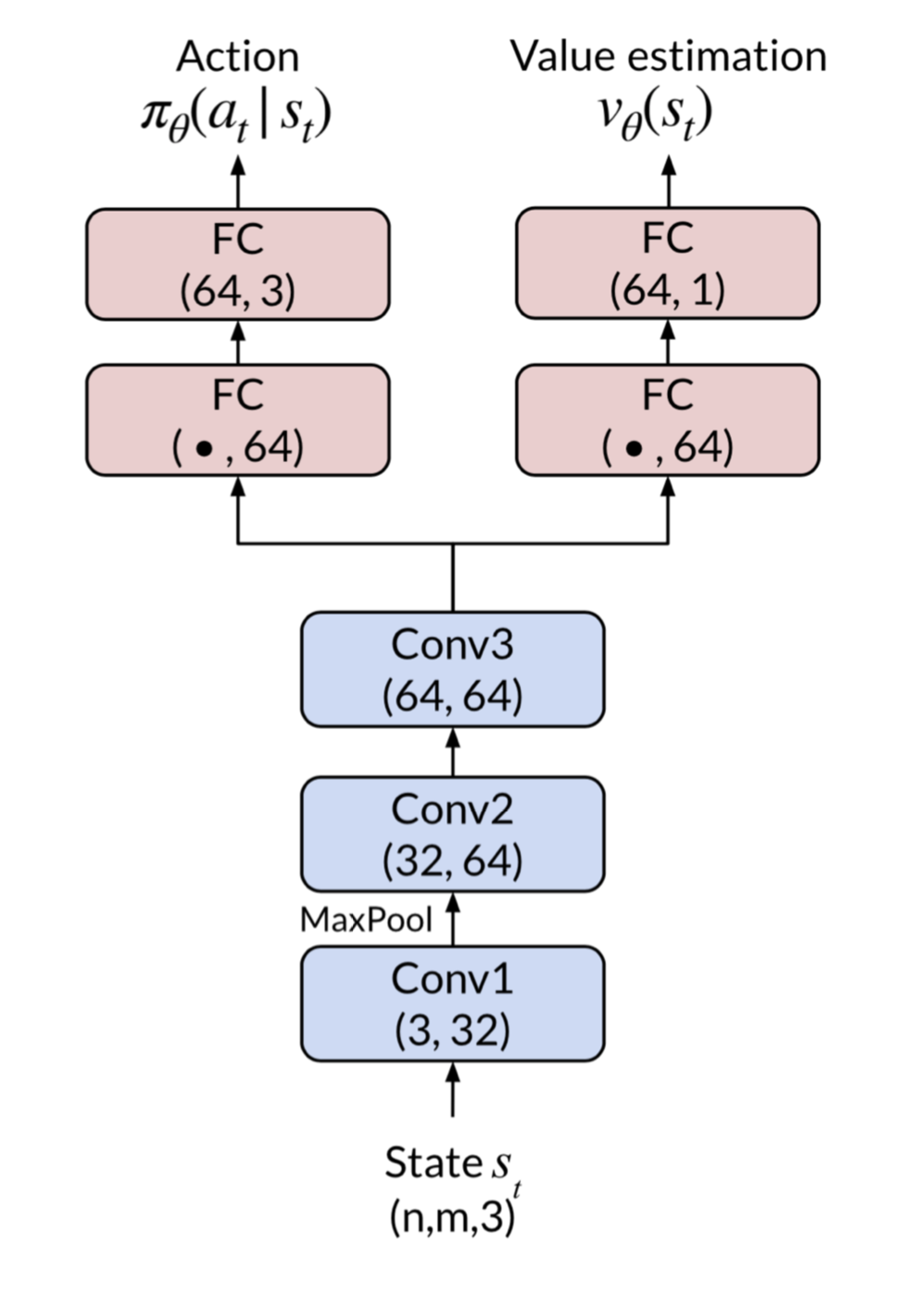}
    \caption{Architecture of the actor-critic policies on MiniGrid. Conv is Convolutional Layer and filter sized is described below.  FC is a fully connected layer.}
    \label{fig:my_label}
\end{figure}

For experiments, we used the PPOC algorithm with parallelized data collection and GAE.  0.1 is the optimal deliberation cost. Each environment is run with 10 random network initialization.  As before, in Table \ref{mini-params}, we show some of the policy optimization parameters for MiniGrid Tasks.  We rely on an actor-critic network architecture for these tasks.  Since the state space is relatively large and spatial features are relevant, we use 3 convolutional layers in the network.  The network architecture is detailed in Figure \ref{fig:my_label}. $n$ and $m$ are defined by the grid dimensions.

The discriminator network is again an neural network with the optimal architecture of 3 linear layers of 150 hidden states, each with ReLU activation followed by a single node linear layer for output.

\begin{table}[tbh]
    \caption{Policy optimization parameters for benchmark tasks in MiniGrid}
    \label{mini-params}
  \begin{center}
    \begin{tabular}{lr}
      \toprule
      Parameter & Value  \\
      \midrule
      Adam optimizer learning rate                                  & $7\cdot 10^{-4}$      \\ 
      Adam $\epsilon$  & $10^{-5}$\\
      entropy coefficient
      & $10^{-2}$\\
      value loss coefficient & 0.5\\
            discount                                        & 0.99      \\

      maximum norm of gradient in PPO &0.5\\
      number of PPO epochs &4\\
      batch size for PPO &256\\
      entropy coefficient                    & $10^{-2}$      \\
      clip parameter &0.2
             \\
    
      \bottomrule
    \end{tabular}
  \end{center}
\end{table}

\end{document}